\documentclass[10pt,twocolumn,letterpaper]{article}

\usepackage{iccv}
\usepackage{times}
\usepackage{epsfig}
\usepackage{graphicx}
\usepackage{amsmath}
\usepackage{amssymb}
\usepackage{booktabs}
\usepackage{enumitem}
\usepackage{multirow}
\usepackage{multicol}
\usepackage{pifont}
\newcommand{\cmark}{\ding{51}}%
\newcommand{\xmark}{\ding{55}}%
\usepackage{romannum}
\usepackage{subcaption}
\usepackage{wrapfig,lipsum,booktabs}
\usepackage{indentfirst}
\usepackage[accsupp]{axessibility} 
\usepackage{tikz}
\usetikzlibrary{fit}
\usetikzlibrary{calc,shapes}
\usetikzlibrary{decorations.pathmorphing} 
\usetikzlibrary{fit}					
\usetikzlibrary{backgrounds}	
\usetikzlibrary{pgfplots.groupplots}
\usepackage[utf8]{inputenc}
\usepackage{pgfplots}
\usepgfplotslibrary{groupplots,dateplot}

\usepackage[pagebackref=true,breaklinks=true,letterpaper=true,colorlinks,bookmarks=false]{hyperref}
\usepackage{longtable}
\iccvfinalcopy 

\def\iccvPaperID{12685} 

\ificcvfinal\fi

\begin{document}

\newcommand{\revise}[1]{{\color{blue}{#1}}}

\newcommand{\sm}[1]{{\color{blue}{\small\bf\sf [Shuang: #1]}}}  
\newcommand{\yao}[1]{{\color{red}{\small\bf\sf [yao: #1]}}}  
\newcommand{\ys}[1]{{\color{green}{\small\bf\sf [Yanchao: #1]}}}
\newcommand{\rj}[1]{{\color{brown}{\small\bf\sf [Ruijie: #1]}}}

\newcommand{\ours}{{DualMind}\xspace}
\newcommand{\loss}{\mathcal{L}}

\newcommand{\tasks}{\mathcal{T}}
\newcommand{\mdp}{\mathcal{M}}
\newcommand{\states}{\mathcal{S}}
\newcommand{\actions}{\mathcal{A}}
\newcommand{\obss}{\mathcal{O}}
\newcommand{\transition}{P}
\newcommand{\reward}{R}
\newcommand{\emission}{E}

\newcommand{\history}{h}
\newcommand{\hisspace}{\mathcal{H}}

\newcommand{\rep}{\phi}

\newcommand{\fwdmath}{\text{fwd}}
\newcommand{\invmath}{\text{inv}}
\newcommand{\randinvmath}{\text{mask-inv}}

\newcommand{\ourmod}{{Enc-Dec Control Transformer}\xspace}
\newcommand{\xatten}{{XAtten.}\xspace}
\newcommand{\duelph}{{Dual-phase}\xspace}

\definecolor{darkgray176}{RGB}{176,176,176}
\definecolor{darkgreen}{RGB}{0,100,0}
\definecolor{darkorange}{RGB}{255,140,0}
\definecolor{lightgray204}{RGB}{204,204,204}
\definecolor{silver}{RGB}{192,192,192}
\definecolor{steelblue}{RGB}{70,130,180}
\definecolor{spink}{RGB}{255,172,175}
\definecolor{sblue}{RGB}{0.596078431372549,0.556862745098039,0.835294117647059}

\definecolor{ours}{rgb}{0.886274509803922,0.290196078431373,0.2}
\definecolor{jointly}{rgb}{0.203921568627451,0.541176470588235,0.741176470588235}
\definecolor{il}{rgb}{0.596078431372549,0.556862745098039,0.835294117647059}
\definecolor{gato}{rgb}{0.984313725490196,0.756862745098039,0.368627450980392}
\definecolor{gato2}{rgb}{0.456862745098039,0.629411764705882,0.158823529411765}

\newenvironment{rebuttal}{\par\color{blue}}{\par}

\title{Is Imitation All You Need? Generalized Decision-Making with Dual-Phase Training}

\author{
\textbf{Yao Wei$^{^{2,6}}$},\textbf{Yanchao Sun$^{^3}$}, \textbf{Ruijie Zheng$^3$}, \textbf{Sai Vemprala$^5$}, \textbf{Rogerio Bonatti$^{^1}$}, \textbf{Shuhang Chen $^5$}
\\\hspace{0.2em}
\textbf{Ratnesh Madaan$^{^1}$}, \textbf{Zhongjie Ba$^{^{2,6}}$}, \textbf{Ashish Kapoor$^{^5}$} and \textbf{Shuang Ma$^{^1{^\star}}$}\\
{ {}$^1$Microsoft}
{ {}$^2$Zhejiang University}
{ {}$^3$University of Maryland ,College Park } \\
{ {}$^5$ Scaled Foundations} { {}$^6$Jiaxing Research Institute, Zhejiang University}\\
{ \small {}$^\star$ Project lead. Corresponding to} 
{\tt\small yunyikristy@gmail.com} 
}



\maketitle
\ificcvfinal\thispagestyle{empty}\fi
\pagenumbering{arabic}

\begin{abstract}
    We introduce \ours, a generalist agent designed to tackle various decision-making tasks that addresses challenges posed by current methods, such as overfitting behaviors and dependence on task-specific fine-tuning. \ours uses a novel ``Dual-phase" training strategy that emulates how humans learn to act in the world. The model first learns fundamental common knowledge through a self-supervised objective tailored for control tasks and then learns how to make decisions based on different contexts through imitating behaviors conditioned on given prompts. \ours can handle tasks across domains, scenes, and embodiments using just a single set of model weights and can execute zero-shot prompting without requiring task-specific fine-tuning. We evaluate \ours on MetaWorld~\cite{yu2020meta} and Habitat~\cite{ramakrishnan2021hm3d:habitat_benchmark} through extensive experiments and demonstrate its superior generalizability compared to previous techniques, outperforming other generalist agents by over 50$\%$ and 70$\%$ on Habitat and MetaWorld, respectively. On the 45 tasks in MetaWorld, \ours achieves over 30 tasks at a 90$\%$ success rate. Our source code is available at \href{https://github.com/yunyikristy/DualMind}{https://github.com/yunyikristy/DualMind}.
\end{abstract}



\section{Introduction}
Transformer-based models, combined with large-scale data, have shown success in generalizing across various tasks in both language and vision. 
Notable examples include BERT~\cite{devlin2019bert}, GPT~\cite{radford2018improving:gpt}, MAE~\cite{he2022masked:mae}, CLIP~\cite{radford2021learning:clip} and Flamingo~\cite{alayrac2022flamingo}, etc. 
Recently, there has been a significant focus on developing such general-purpose models for sequential decision-making and control tasks, such as GATO~\cite{reed2022generalist}. The prominent approach is to train a decoder-only Transformer with Imitation Learning (IL) on massive datasets from all targeted tasks. By training with prompts, the model can perform zero-shot inference with just task prompts. 

 \begin{figure}
    \centering
    \includegraphics[width=.475\textwidth]{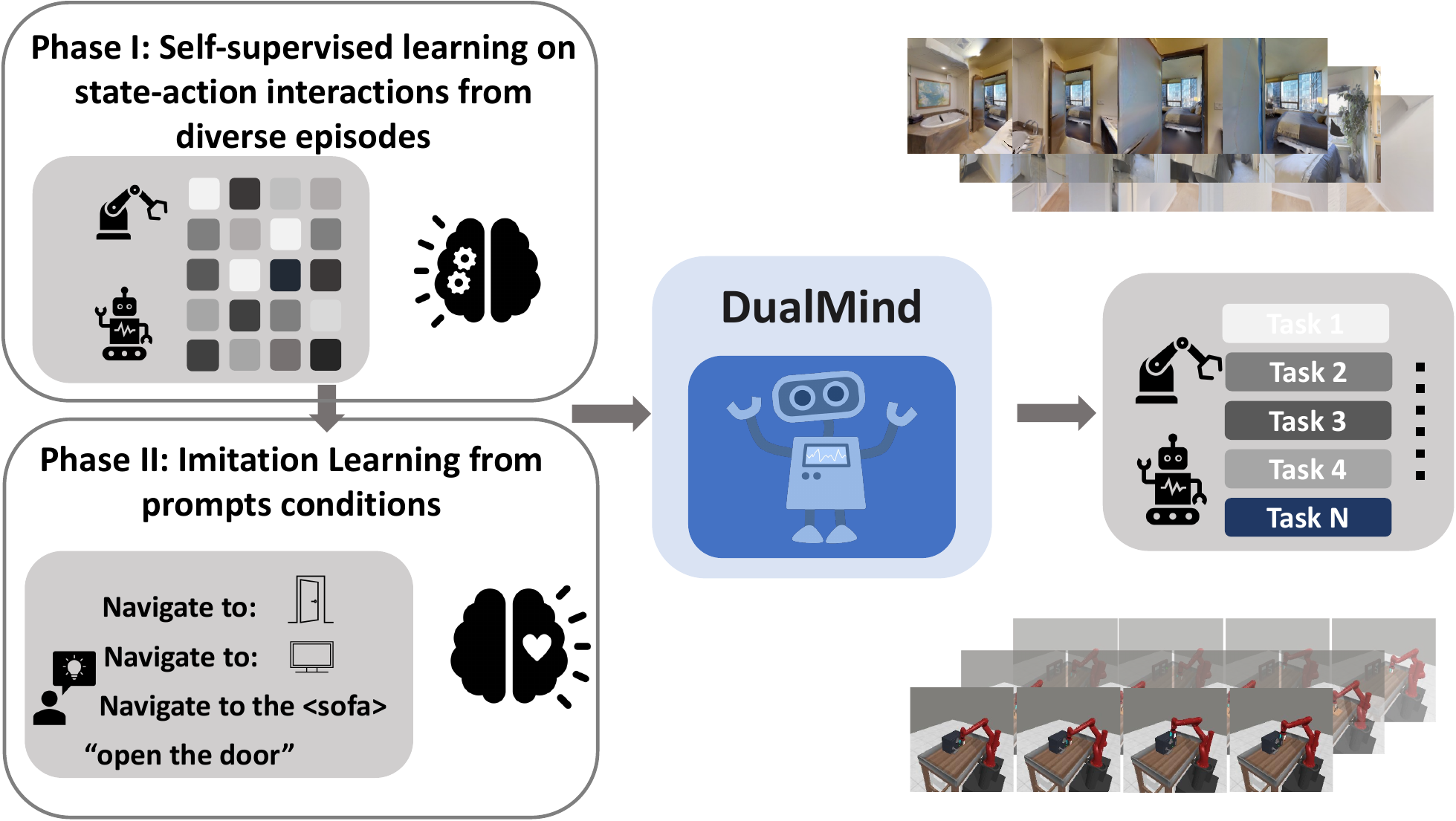}
    \caption{A high-level overview of \ours's \duelph training scheme. 
}
    \label{fig:my_label}
\end{figure}

However, such IL-based approaches to general-purpose models face limitations when it comes to sequential control tasks, as highlighted below:
(1) \textit{Memorizing behaviors hinders generalization to diverse tasks}: 
Imitating expert behaviors can lead to memorization and over-fitting of specific behaviors that may not be applicable to new situations or variations of tasks, thus limiting the model's ability to generalize. This limitation is particularly challenging when dealing with a wide range of decision-making tasks that have vastly different configurations, transition functions, and state and action spaces.
(2) \textit{Dependence on high-quality data impedes practical application}:
IL methods rely heavily on the availability of high-quality expert demonstrations, which can be difficult and expensive to obtain. 
When the available data is of low quality or not representative of the target task, the performance of the model may suffer. 

In light of the aforementioned limitations, self-supervised pretraining has emerged as a viable solution. By focusing on learning common underlying information, a pretrained model can be better equipped to handle diverse tasks. Recently, a study known as SMART~\cite{sun2023smart} has demonstrated the potential of self-supervised pretraining for multi-task decision-making. 

Although SMART has shown promising results in promoting generalization, it still requires additional fine-tuning to adapt to each task. Furthermore, it has only been demonstrated on a small set of tasks on Deepmind control suite (DMC)~\cite{tassa2018deepmind:dmc}.
For decision-making problems that involve numerous tasks with different configurations, finetuning the model for each task can become time-consuming and resource-intensive.

Given the limitations of both IL and self-supervised pretraining discussed earlier, a natural question arises: \textit{How can we develop a decision-making approach that achieves a high degree of generalization without requiring task-specific fine-tuning?} In this paper, we propose \ours, a generalist agent, to address this question, which stands for our proposed \duelph training scheme. 
The name `DualMind' is derived from our main idea of \duelph training for generalized decision-making.
Our approach introduces an Encoder-Decoder Control Transformer (\ourmod) that models state-action interactions from complex high-dimensional observations. 
To further improve computational efficiency, \ours uses TokenLearner~\cite{NEURIPS2021_6a30e32e:tokenlearner} as an attention-based Information Bottleneck (IB)~\cite{tishby2000information:IB} to compress the number of tokens so that to speed up training and inference.
Building upon \ourmod, we propose a \duelph training scheme that initially prioritizes policy-independent transition probabilities and encourages the model to capture both short- and long-term temporal granularities. To facilitate zero-shot prompting, we train a second phase on a small fraction of model parameters to learn a generic policy by conditioning on various prompts (such as images, annotations, and language instructions) using a cross-attention mechanism (\xatten). 
The \duelph training scheme parallels how humans learn to act in the world by first learning underlying common knowledge and subsequently making decisions based on different contexts.
Our contributions are summarized below:
\begin{enumerate}[noitemsep,leftmargin=*,topsep=0pt]
\setlist{nolistsep}
    \item We propose \ours, a solution for general-purpose decision-making that can handle various tasks using a single set of weights without task-specific fine-tuning.
    \item We introduce a \duelph training scheme that overcomes limitations of IL and self-supervised learning.
    \item We propose an Encoder-Decoder Transformer (\ourmod) that efficiently learns state-action transitions from high-dimensional observation spaces.
    \item We conduct extensive experiments on Metaworld~\cite{yu2020meta} and Habitat~\cite{ramakrishnan2021hm3d:habitat_benchmark} and show that \ours outperforms other generalist agents by over 50$\%$ and 70$\%$ on Habitat and MetaWorld, respectively. We also analyze and ablate different design choices to demonstrate the superior generalizability of \ours.
\end{enumerate}

\section{Related work}

\begin{table*}[t!]
    \centering
    \small 
    \begin{tabular}{l|lll}
    \toprule
                                         & Self-superv.                  & IL-prompt                      & \duelph (ours) \\ \hline
        \multirow{2}{*}{Learning}        & Pre.: generic info.     & Cond. generic policy                 & \Romannum{1}: generic info. \\
                                         & FT: task-specific policy      &                                & \Romannum{2}: cond. generic policy \\ \hline
        \multirow{2}{*}{Data}            & Pre: Multi-task large set     & Multi-task large set+prompts            & \Romannum{1}: Multi-task large set \\
                                         & FT: Single-task small set     &                                & \Romannum{2}: +prompts \\ \hline
        \multirow{2}{*}{Optim. weights}  & Pre: whole model              &                     & \Romannum{1}: Entire model \\ 
                                         & FT: Entire/freeze+Task heads & Entire model                    & \Romannum{2}: Partial/freeze+XAtten. \\ \hline
        Inference task                   & Single                        & Multiple                       & Multiple \\ \hline
        No need FT                   & \xmark                        & \cmark                         & \cmark \\ \hline
        Zero-shot promp.              & \xmark                        & \cmark                         & \cmark \\ \hline
        Final utilization                & Many models for each task     & Single model     & Single model \\
        \bottomrule
    \end{tabular}
    \caption{Comparisons of different training approaches.}
    \vspace{-1em}

    \label{tab:study-approach}
\end{table*}

\emph{Pretraining Visual Representations for Policy Learning:} Recent studies such as R3M~\cite{nair2022r3m}, APV~\cite{seo22} VPT~\cite{baker2022video}, NRNS~\cite{hahn2021no}, PVR~\cite{parisi2022unsurprising} and MVP\cite{radosavovic2022real} have shown that pre-trained visual representations can significantly enhance the efficiency of downstream policy learning. However, these works mainly focus on learning object-centric semantics, potentially losing essential control-relevant information. To address this issue, VIP~\cite{ma2023vip} formulates the problem as an offline goal-conditioned RL problem and proposes a visual representation algorithm capable of generating dense reward functions for downstream robotics tasks. On the other hand, COMPASS~\cite{ma2022compass} introduces a general-purpose pretraining pipeline that effectively integrates multimodal signals for autonomous systems.

\emph{Transformer-Based Foundational Model:} The use of high-capacity transformer architectures trained on large-scale datasets has led to significant breakthroughs in various domains. Examples include language models such as BERT~\cite{devlin2019bert}, GPT-3~\cite{Brown2020}, T5~\cite{2020t5}, and PaLM~\cite{chowdhery2022palm}, as well as vision and vision-language models such as MAE~\cite{he2022masked:mae}, Multi-MAE~\cite{bachmann2022multimae}, BiT~\cite{kolesnikov2020}, MuST~\cite{ghiasi21}, Flamingo~\cite{alayrac2022flamingo}, and CLIP~\cite{radford2021learning:clip}. For decision-making tasks, recent work such as SMART~\cite{sun2023smart} has proposed a self-supervised pretraining framework tailored for control tasks. For robotics control problems, PACT \cite{bonatti2022pact} has shown that a pretrained representation could speed up various downstream tasks of mobile agents, such as navigation and localization.

\emph{A General-Purpose Model for Control:} Since the groundbreaking success of GPT~\cite{radford2018improving:gpt}, recent research has focused on using Transformer decoder-based models to tackle control tasks in an auto-regressive manner. Decision Transformer (DT) \cite{chen2021decision, zheng22dt} builds on the architecture of GPT to create a generalist agent for sequential decision-making tasks. This has been followed by Multi-game DT~\cite{lee2022multigame} and Online-DT~\cite{zheng2022online:onlinedt}, which demonstrate the potential of DTs for multi-task and online learning. GATO~\cite{reed2022generalist} imitates expert demonstrations from a vast dataset and showcases its ability to handle a large number of tasks. VIMA~\cite{jiang2022vima} is an agent that can accept multi-modal prompts for solving various robotics manipulation tasks. In real-life applications, RT-1~\cite{brohan2022rt} has demonstrated the efficacy of this approach in robotic control.

\section{Preliminary and Overview of \ours}
\label{sec:prelim}

\begin{table}[t!]
    \centering
    \small 
    \begin{tabular}{c|llllllll}
    \toprule
    Bench.  & Dom.      & Sce.  & Emb.  & Prom.        & Tasks & Epis. \\
    Meta.  & Man. & 1            & 1                & inst. & 50       & 50K \\
    Habit.    & Nav.   & 933          & 1                & Obj. / Img   & 27      & 50K\\ \hline
    Total  & 2            & 934         & 2                & 3              & 77       & 100K \\  
    \bottomrule
    \end{tabular}
    \caption{Dataset summerization Dom.: domains, Sce.: number scenes, Emb.: number of embodiments, Prom.: types of prompts, Epis.: number of episodes.}
    \vspace{-1em}
    \label{tab:data}
\end{table}


\subsection{Problem formulation} \label{ssec:formulation}
We focus on a set of tasks, denoted as $\tasks$, from two representative benchmarks, namely Metaworld~\cite{yu2020meta} and Habitat~\cite{ramakrishnan2021hm3d:habitat_benchmark}, which cover the \textit{Manipulation} and \textit{Navigation} domains, respectively. 
As shown in Table~\ref{tab:data}, our selection of these two benchmarks allows us to conduct a comprehensive study on tasks with a wide variety of characteristics. Here, we define a task as a 
partially observable Markov decision process (POMDP).
The tasks we consider span across several factors, as defined below:

\begin{itemize}[noitemsep,leftmargin=*,topsep=0pt]
\setlist{nolistsep}
\item \textit{Domain}: This refers to tasks with different state/action spaces and application scenarios. In our study, \textit{Manipulation} and \textit{Navigation} are two domains we focus on.
\item \textit{Embodiment}: This factor is used to differentiate tasks that have different physics and action spaces. For instance, a robot arm and an embodied agent in MetaWorld and Habitat are considered as different embodiments. Differences can also exist in the same domain, such as arms with distinct joint torques and/or hardware configurations.
\item \textit{Scene}: This refers to tasks that are performed in different observation spaces, state spaces, and world structures. For example, in Habitat, agents that navigate in different rooms should adapt to various visual appearances and geometry structures.
\item \textit{Prompt}: This factor captures different forms of prompt conditions. In MetaWorld, prompts are natural language instructions, while in Habitat, we use a single RGB image or an object annotation as the navigation goal to prompt our model.
\end{itemize}

\begin{figure*}[t!]
    \centering
    \includegraphics[width=1. \textwidth]{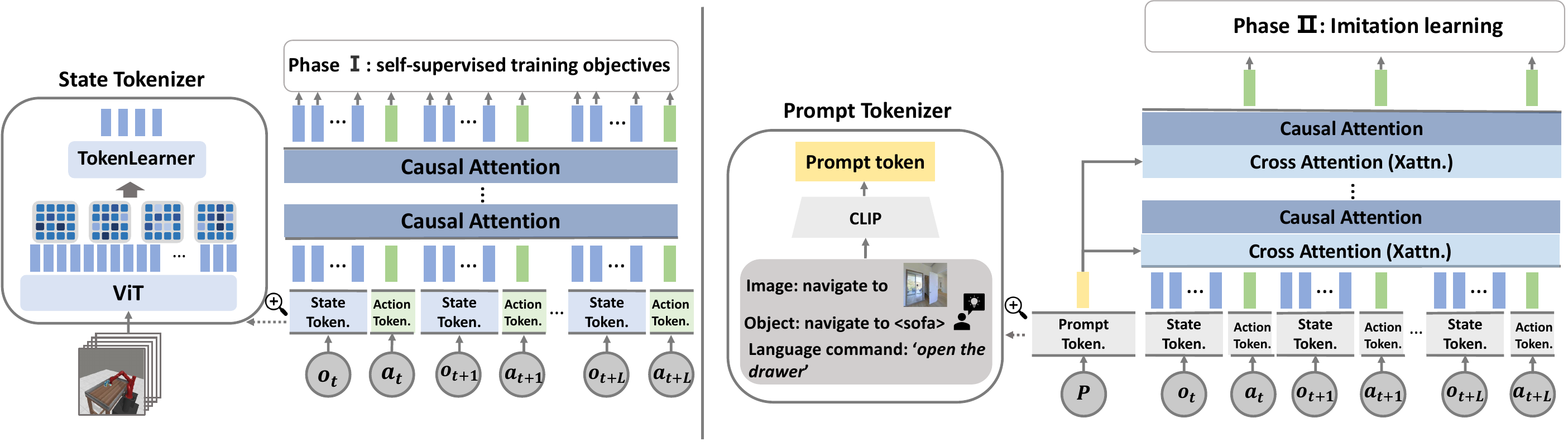}
    \caption{The architecture diagram of \ours. \textbf{Left: Phase \Romannum{1}.} Agent is trained with self-supervised learning objectives. During this phase, Transformer encoder and decoder jointly trained. \textbf{Right: Phase \Romannum{2}.} Agent is trained with prompt conditional imitation learning. We tokenize task prompts with a pretrained CLIP encoder, and condition the Transformer decoder on the prompt through \xatten layers. The gray color indicates frozen modules. (Detailed training objectives are in Sec.~\ref{ssec:obj}.)}
    \label{fig:dual_phase}
\end{figure*}

\subsection{Overview of \duelph training scheme} 
In this section, we provide a brief overview of \ours and compare it with two other prominent approaches: self-supervised pretraining (Self-superv.) and Imitation Learning with prompt conditions (IL-prompt). We also provide insights into the central idea behind our proposed approach. A summarized comparison of these approaches is shown in Table~\ref{tab:study-approach}.
 
As shown in Fig.~\ref{fig:dual_phase}, 
In Phase \Romannum{1}, we train the entire \ourmod (Sec.~\ref{ssec:model}) with a self-supervised training objective to capture generic information of state-action transitions. 
In Phase \Romannum{2}, we train only a small part of \ourmod attached with \xatten on a diverse set of prompts for a conditional generic policy. After the \duelph training, we can obtain one model with a single set of weights that can be directly applied to a large number of tasks with corresponding prompts.


Compared to other generalist agents like GATO~\cite{reed2022generalist}, which trains an imitating policy directly, \ours demonstrates superior generalization capability. Moreover, our Phase \Romannum{2} requires training only a small fraction of model weights while freezing the remaining parts, resulting in faster learning and reduced training cost when optimizing the model with the same number of iterations. Additionally, compared to self-supervised learning approaches such as SMART~\cite{sun2023smart}, \ours is simple and effective, making it suitable for a wide range of application scenarios.
\subsection{Insights}

The central idea behind our \duelph is to mimic how humans learn to act in the world, first by learning underlying common knowledge and then by learning to make decisions based on different contexts. 
Our approach relates to InstructGPT~\cite{ouyang2022training}, which aims to align language models with user intent by fine-tuning them with human feedback.
In analogy to InstructGPT, our Phase \Romannum{1} can be considered as learning a general model that captures the common essential information. However, as stated in InstructGPT, this is different from the objective of ``following task instructions (i.e. prompt conditions)," and thus such a model is \textit{misaligned}. Therefore, in the second phase, we leverage conditional IL to align the model so that it can perform well for any given prompts.

\section{Approach}


In this section, we introduce our proposed \ours.
We present the model architecture in Section~\ref{ssec:model}, and illustrate the training objective for \ours in Section~\ref{ssec:obj}. 

\subsection{Model Architecture}
\label{ssec:model}
We propose an Encoder-Decoder Control Transformer to process state-action interaction sequences, as illustrated in Figure \ref{fig:dual_phase}. 
The implementation details of each component in the \ourmod are outlined below.

\textbf{State tokenizer.} We utilize a ViT model~\cite{dosovitskiy2020image} to tokenize raw pixel states. To reduce the computational burden of dealing with sequential decision-making tasks, we leverage an attention-based Information Bottleneck (IB) to further compress the number of tokens so as to speed up training and inference (Fig.~\ref{fig:dual_phase}-left). Specifically, we use TokenLearner~\cite{NEURIPS2021_6a30e32e:tokenlearner} which is an element-wise attention module that learns to soft-select image tokens, passing only the important ones to subsequent layers. The inclusion of TokenLearner sub-samples the 196 state tokens that come out of ViT to just 8 tokens that are then passed to the Transformer decoder layers.

\textbf{Action tokenizer.} To handle both continuous and discrete action spaces in our two domains, we adapt a strategy similar to GATO~\cite{reed2022generalist} by discretizing continuous actions into bins. We first flatten the actions into sequences of floating point values in row-major order, and then 
discretizing them into 256 uniform bins. Discrete actions are tokenized into 256 bins in the same way.
 
\textbf{Transformer decoder.} Our transformer decoder architecture is similar to Control Transformer~\cite{sun2023smart}, but with a modification. In our approach, we encode each state into 8 tokens, which is different from SMART's single token representation. This modification enables richer representation learning, making it suitable for more complex visual control environments.

\textbf{Prompt tokenizer.} We tokenize prompts using a pretrained CLIP encoder~\cite{radford2021learning:clip}. For ``image goal" prompts in Habitat, we use the CLIP image encoder, while for ``object goal" prompts in Habitat and ``language instruction" prompts in MetaWorld, we use the CLIP text encoder. A learnable linear layer is added on top of the CLIP encoders to map all prompts to prompt tokens with the same dimensions. During training in both phases, we freeze the CLIP encoders.

\textbf{\xatten layer.} We condition the Transformer decoder by training it to learn from the prompt sequence through a series of cross-attention layers. The output sequence from each cross-attention layer is computed by softmax$(\frac{q_{H}k_p^T}{\sqrt{d}})v_P$, where $H$ is the sequence of episodes, $P$ is prompt, and $d$ is the embedding dimension. This design builds a stronger connection between the prompts and the demonstrations, which is an improvement over prefix-style prompting approaches~\cite{reed2022generalist}. We will show the benefits of this design in Sec.~\ref{ssec:ablate-prompt}.


\subsection{Training objectives}
\label{ssec:obj}
\begin{figure}
    \centering
    \includegraphics[width=.5\textwidth]{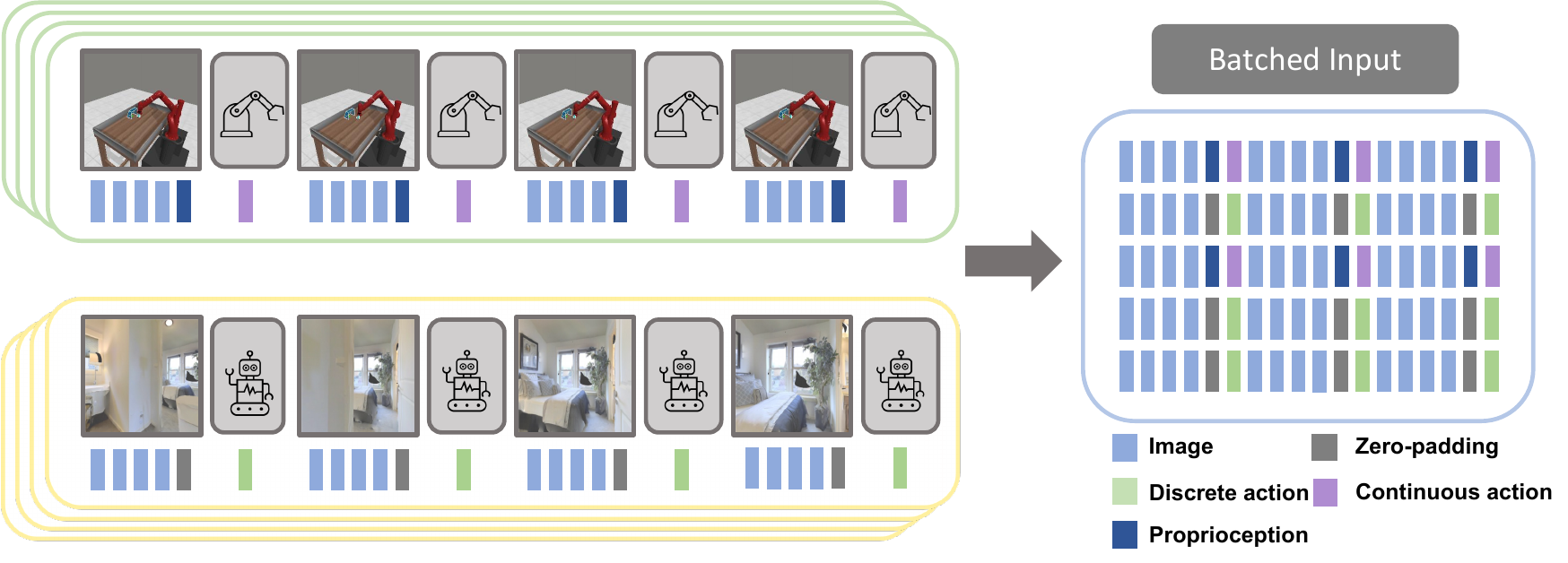}
    \caption{Batch input when training on multiple domains.}
    \label{fig:batch_input}
\end{figure}


\textbf{Phase \Romannum{1}: self-supervised SMART training.}
The goal of this phase is to learn a good representation that captures control-relevant information shared across tasks.
In this phase, we jointly train the encoder and the decoder following the self-supervised training objectives of SMART~\cite{sun2023smart}. 
We use $F_\theta$ to denote the learned model with parameterization $\theta$, such that $F_\theta(o_{i:j},a_{i:j})$ refers to the output tokens of the decoder corresponding to raw inputs $o_{i:j}$ and $a_{i:j}$, the observation and action sequence from step $i$ to step $j$. 
For a sequence of observations and actions denoted as $\{ o_t, a_t, \cdots, o_{t+L}, a_{t+L}\}$ with context length $L$, we minimize the following objective.
{\small
\begin{align}
    \loss_{P1} &:= \loss_1 + \loss_2 + \loss_3, \text{ where} & \\
    \loss_1 &:= \sum\nolimits_{i=0}^{L-1} l\left( f_1(F_\theta(o_{t:t+i}, a_{t:t+i})), \bar{\phi}({o}_{t+i+1}) \right), & \\
    \loss_2 & := \sum\nolimits_{i=1}^{L} l\left( f_2(F_\theta(o_{t:t+i+1}, a_{t:t+i-1} ), a_{t+i} \right), &\\
    \loss_3 & := \sum\nolimits_{i=1}^{L-1} l\left( f_3(F_\theta(\mathsf{Mask}(o_{t:t+L},a_{t:t+L})), a_{t+i} \right).
\end{align}
}%
Here $l$ is a loss function that is selected by the variable type. For latent states, we use a mean squared error, while for discrete actions, we use the cross-entropy loss. $\loss_1$ is to learn a forward prediction head $f_1$ that can predict the next state representation based on the historical interactions. Since the groundtruth state representation is unknown, we use the learned state embedding from the ViT model to encode the next observation, denoted as $\bar{\phi}$ where the overline stands for gradient stopping.  
$\loss_2$ aims to recover the action token in each step conditioning on the history and the next state. $\loss_3$ masks a proportion of input tokens and learns to recover the masked actions, which can extract long-term temporal dependence for control. 

\vspace{0.7em}
\textbf{Phase \Romannum{2}: Imitation learning with prompt conditions.} 
In this phase, we train the model to follow prompt conditions. 
We formulate various tasks as a conditional generation problems, where the conditions can be goals, commands, prompts, etc. 
During Phase \Romannum{2}, we let the agent learn a conditional policy, using expert trajectories with associated prompts. 
Let $\psi$ be the prompt tokenizer, and $\pi$ be the learned policy whose inputs are the representation tokens given by the decoder. 
For an expert sequence $\{ o_t, a_t, \cdots, o_{t+L}, a_{t+L}\}$ with prompt $P$, we minimize loss
{\small{
\begin{equation}
    \loss_{P2} := \sum\nolimits_{i=0}^{L-1} l(\pi(F_\theta(o_{t:t+i}, a_{t:t+i}; \psi(P))), a_{t+i+1}).
\end{equation}
}}%
Note that in this phase, we do not train the entire model $F_\theta$, and instead only re-train a small fraction of it. More discussion is in Sec.~\ref{ssec:ablate-phase}. 


\section{Experiments}

\subsection{Experimental setup}
\label{ssec:exp_setup}

\textbf{Data.}
We evaluate and train \ours on two benchmarks, Habitat~\cite{ramakrishnan2021hm3d:habitat_benchmark} and MetaWorld~\cite{yu2020meta}. Habitat is a photorealistic simulation platform for research in Embodied AI, emphasizing active perception and long-term planning, ·while MetaWorld is a simulated benchmark for multi-task learning and meta-reinforcement learning, comprising 50 distinct robotic manipulation environments. Training on datasets collected from both these benchmarks allows us to demonstrate the model's generalizability across domains, embodiments, scenes, and prompts. We provide a detailed introduction to these factors in Sec.~\ref{ssec:formulation} and summarize them in Table~\ref{tab:data}. Additionally, we use 10 tasks as an out-of-distribution testbed to showcase the model's generalization capability. More details about our data collection process can be found in Appendix~\ref{app:imp-detail}.

\textbf{Comparing baselines.}{\label{ssec:comp-baseline}}
We compare \ours with existing transformer-based approaches and present results from two versions of our model: a generalist agent trained on the full dataset (\texttt{\ours}) and a single-domain specialist trained only on data from either MetaWorld or Habitat (\texttt{\ours/single}). To ensure fair comparisons, we implemented related works ourselves and trained and evaluated them on the same data and model architecture. We provide information on each baseline below:
\begin{itemize}[noitemsep,leftmargin=*,topsep=0pt]
\setlist{nolistsep}
    \item \texttt{IL-only} is a model trained only through prompt-conditioned imitation learning, which is related to GATO but uses a different prompting conditioning method.
    \item \texttt{SMART-only} is a model trained only using SMART training objectives (purely self-supervised).
    \item \texttt{Jointly} is a model jointly trained with both SMART objectives and prompt-conditioned Imitation Learning loss.
    \item \texttt{GATO*} is the model described in the original paper. We include its reported performance on the Metaworld benchmark for reference. Notably, this model has 1.18 billion parameters and was trained on massive datasets, including 94.6k episodes from Metaworld. In comparison, \texttt{\ours} has 175 million parameters and was trained on a smaller dataset consisting of 100k episodes, of which 50k are from MetaWorld.
    
    \item \texttt{GATO-CT} is a model we implemented ourselves, reproducing the main technical approaches presented in the original paper. For a fair comparison, we used the same base model architecture (\ourmod), but replaced our \xatten-based prompting approach with their proposed prefix prompting approach. Similar to \texttt{IL-only}, only imitation learning loss is used to train with prompt conditions. We provide a detail description in Appendix.~\ref{app:imp-detail}.

\end{itemize}

\vspace{.5em}
\textbf{Implementation details.}
Our implementation of \ours uses a Transformer-based architecture consisting of a ViT-B~\cite{gupta2022maskvit} model, a TokenLearner~\cite{NEURIPS2021_6a30e32e:tokenlearner}, and a GPT model~\cite{radford2018improving:gpt} as the encoder and decoder, respectively. The decoder consists of 8 layers and 8 attention heads, with a context length of L=6 and an embedding size of d=512\footnote{We found that longer context lengths can produce better performance, particularly on tasks that rely on long-range temporal dependencies. See Appendix~\ref{app:more-exp} for more ablations.}. 
We trained our model with the AdamW optimizer~\cite{loshchilovdecoupled} and a learning rate of 5e-5 for both training phases. In Phase \romannum{1}, we trained the model for around 40 hours with BS=16 on 5x8xV100 GPUs. In Phase \romannum{2}, the model was trained for about 12 hours with BS=128 on 2x8xV100 GPUs. Further implementation details are provided in Appendix~\ref{app:imp-detail}.


\subsection{Capabilities of \ours} {\label{sec:capabilities}}


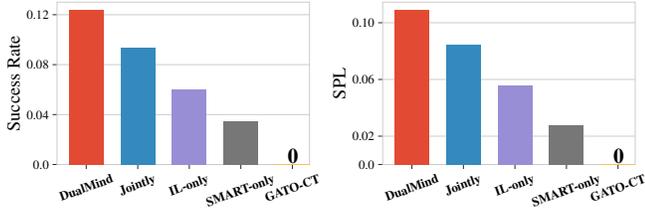
\begin{figure}[t!]
    \centering
    \begin{subfigure}[t]{0.47\columnwidth}
        \centering
        \resizebox{1.1\columnwidth}{!}{
\begin{tikzpicture}

\begin{axis}[
width=10cm,
height=7cm,
axis line style={white!80!black},
tick align=outside,
tick pos=left,
x grid style={white!80!black},
xmin=0.065, xmax=0.8,
xtick style={color=white!15!black},
xtick={0.15,0.3,0.45,0.6,0.75},
xticklabel style={rotate=20, font=\fontsize{12}{6}\selectfont},
xticklabels={\textbf{\ours},\textbf{Jointly},\textbf{IL-only},\textbf{SMART-only},\textbf{GATO-CT}},
y grid style={white!80!black},
ylabel={\LARGE{Success Rate}},
ylabel style={yshift=6pt},
yticklabel style={ font=\fontsize{12}{6}\selectfont},
ymajorgrids,
ymin=0, ymax=0.130095,
scaled y ticks=false,
ytick style={color=white!15!black},
ytick={0.0,0.04,0.08,0.12},
yticklabels={0.0,0.04,0.08,0.12},
]
\draw[draw=none,fill=ours] (axis cs:0.1,0) rectangle (axis cs:0.2,0.1239);
\draw[draw=none,fill=jointly] (axis cs:0.25,0) rectangle (axis cs:0.35,0.0935);
\draw[draw=none,fill=il] (axis cs:0.4,0) rectangle (axis cs:0.5,0.06);
\draw[draw=none,fill=white!46.6666666666667!black] (axis cs:0.55,0) rectangle (axis cs:0.65,0.0349);
\draw[draw=none,fill=gato] (axis cs:0.7,0) rectangle (axis cs:0.8,0);
\draw (axis cs:0.715,0.001) node[
  scale=2,
  anchor=base west,
  text=black,
  rotate=0.0
]{\bfseries 0};
\end{axis}

\end{tikzpicture}} 
        \label{sfig:hat_sr}
    \end{subfigure}
    \hfill
    \begin{subfigure}[t]{0.47\columnwidth}
        \centering
        \resizebox{1.1\columnwidth}{!}{
\begin{tikzpicture}

\begin{axis}[
width=10cm,
height=7cm,
axis line style={white!80!black},
tick align=outside,
tick pos=left,
x grid style={white!80!black},
xmin=0.065, xmax=0.8,
xtick style={color=white!15!black},
xtick={0.15,0.3,0.45,0.6,0.75},
xticklabel style={rotate=20, font=\fontsize{12}{6}\selectfont},
xticklabels={\textbf{\ours},\textbf{Jointly},\textbf{IL-only},\textbf{SMART-only},\textbf{GATO-CT}},
y grid style={white!80!black},
ylabel={\LARGE{SPL}},
ylabel style={yshift=6pt},
ymajorgrids,
ymin=0, ymax=0.114555,
ytick style={color=white!15!black},
yticklabel style={ font=\fontsize{12}{6}\selectfont},
scaled y ticks=false,
ytick={0.0,0.02,0.06,0.10},
yticklabels={0.0,0.02,0.06,0.10},
]
\draw[draw=none,fill=ours] (axis cs:0.1,0) rectangle (axis cs:0.2,0.1091);
\addlegendimage{ybar,ybar legend,draw=none,fill=ours}

\draw[draw=none,fill=jointly] (axis cs:0.25,0) rectangle (axis cs:0.35,0.0846);
\addlegendimage{ybar,ybar legend,draw=none,fill=jointly}

\draw[draw=none,fill=il] (axis cs:0.4,0) rectangle (axis cs:0.5,0.0555);
\addlegendimage{ybar,ybar legend,draw=none,fill=il}

\draw[draw=none,fill=white!46.6666666666667!black] (axis cs:0.55,0) rectangle (axis cs:0.65,0.028);
\addlegendimage{ybar,ybar legend,draw=none,fill=white!46.6666666666667!black}

\draw[draw=none,fill=gato] (axis cs:0.7,0) rectangle (axis cs:0.8,0);
\addlegendimage{ybar,ybar legend,draw=none,fill=gato}

\draw (axis cs:0.715,0.001) node[
  scale=2,
  anchor=base west,
  text=black,
  rotate=0.0
]{\bfseries 0};
\end{axis}

\end{tikzpicture}}
        \label{sfig:hat_spl}
    \end{subfigure}
    \vspace{-1.5em}
    \caption{Comparisons of \textbf{generalist agents} on \textit{Habitat 4 scenes} with 3 difficulty levels per scene. We roll out the agents 3 times on each scene and average the defined scores, and compare agents by Success Rate (SR) (left) and Success weighted by Path Length (SPL) (right).}
    \label{fig:g-hab}
\end{figure}
\begin{figure}[t!]
    \centering
    \begin{subfigure}[t]{0.47\columnwidth}
        \centering
        \resizebox{1.1\columnwidth}{!}{
\begin{tikzpicture}

\definecolor{color0}{rgb}{0.886274509803922,0.290196078431373,0.2}
\definecolor{color1}{rgb}{0.203921568627451,0.541176470588235,0.741176470588235}
\definecolor{color2}{rgb}{0.596078431372549,0.556862745098039,0.835294117647059}
\definecolor{color3}{rgb}{0.984313725490196,0.756862745098039,0.368627450980392}
\definecolor{color4}{rgb}{0.456862745098039,0.629411764705882,0.158823529411765}

\begin{axis}[
axis line style={white!80!black},
legend cell align={left},
legend style={
  fill opacity=0.8,
  draw opacity=1,
  text opacity=1,
  at={(0.62,0.53)},
  anchor=center,
  draw=none,
  fill=none
},
tick align=outside,
tick pos=left,
x grid style={white!80!black},
xlabel={\Large{Expert Score/\%}},
xmajorgrids,
xmin=-5, xmax=105,
xtick style={color=white!15!black},
y grid style={white!80!black},
y label style={at={(axis description cs:0.08,0.5)},anchor=south},
ylabel={\large{Number of Tasks}},
ymajorgrids,
ymin=-0.15, ymax=47.15,
ytick style={color=white!15!black}
]
\addplot [ultra thick, color0, mark=*, mark size=3, mark options={solid}]
table {%
0 45
10 45
20 43
30 41
40 39
50 39
60 37
70 37
80 32
90 27
100 18
};
\addlegendentry{\ours}
\addplot [ultra thick, color1, mark=*, mark size=3, mark options={solid}]
table {%
0 45
10 27
20 22
30 19
40 14
50 14
60 14
70 12
80 7
90 5
100 5
};
\addlegendentry{Jointly}
\addplot [ultra thick, color2, mark=*, mark size=3, mark options={solid}]
table {%
0 45
10 28
20 23
30 18
40 15
50 13
60 12
70 10
80 9
90 7
100 5
};
\addlegendentry{IL-only}
\addplot [ultra thick, white!46.6666666666667!black, mark=*, mark size=3, mark options={solid}]
table {%
0 45
10 25
20 18
30 14
40 10
50 9
60 6
70 4
80 3
90 3
100 2
};
\addlegendentry{SMART-only}
\addplot [ultra thick, color3, mark=*, mark size=3, mark options={solid}]
table {%
0 45
10 13
20 9
30 6
40 4
50 3
60 3
70 3
80 3
90 2
100 2
};
\addlegendentry{GATO-CT}
\addplot [ultra thick, color4, dashed, mark=*, mark size=3, mark options={solid}]
table {%
0 45
50 44
80 35
90 3
};
\addlegendentry{GATO*}
\end{axis}

\end{tikzpicture}} 
        \label{sfig:meta45_es}
    \end{subfigure}
    \hfill
    \begin{subfigure}[t]{0.47\columnwidth}
        \centering
        \resizebox{1.1\columnwidth}{!}{
\begin{tikzpicture}

\definecolor{color0}{rgb}{0.886274509803922,0.290196078431373,0.2}
\definecolor{color1}{rgb}{0.203921568627451,0.541176470588235,0.741176470588235}
\definecolor{color2}{rgb}{0.596078431372549,0.556862745098039,0.835294117647059}
\definecolor{color3}{rgb}{0.984313725490196,0.756862745098039,0.368627450980392}

\begin{axis}[
axis line style={white!80!black},
legend cell align={left},
legend style={
  fill opacity=0.8,
  draw opacity=1,
  text opacity=1,
  at={(0.65,0.5)},
  anchor=center,
  draw=none, 
  fill=none
},
tick align=outside,
tick pos=left,
x grid style={white!80!black},
xlabel={\Large{Success Rate}},
xmajorgrids,
xmin=-5, xmax=105,
xtick style={color=white!15!black},
y grid style={white!80!black},
y label style={at={(axis description cs:0.08,0.5)},anchor=south},
ylabel={\large{Number of Tasks}},
ymajorgrids,
ymin=-2.25, ymax=47.25,
ytick style={color=white!15!black}
]
\addplot [ultra thick, color0, mark=*, mark size=3, mark options={solid}]
table {%
0 45
10 42
20 40
30 40
40 37
50 37
60 36
70 36
80 34
90 31
100 28
};
\addlegendentry{\ours}
\addplot [ultra thick, color1, mark=*, mark size=3, mark options={solid}]
table {%
0 45
10 20
20 12
30 12
40 12
50 11
60 8
70 7
80 3
90 1
100 1
};
\addlegendentry{Jointly}
\addplot [ultra thick, color2, mark=*, mark size=3, mark options={solid}]
table {%
0 45
10 17
20 15
30 12
40 9
50 8
60 8
70 6
80 5
90 5
100 3
};
\addlegendentry{IL-only}
\addplot [ultra thick, white!46.6666666666667!black, mark=*, mark size=3, mark options={solid}]
table {%
0 45
10 1
20 1
30 1
40 1
50 1
60 1
70 0
80 0
90 0
100 0
};
\addlegendentry{SMART-only}
\addplot [ultra thick, color3, mark=*, mark size=3, mark options={solid}]
table {%
0 45
10 10
20 7
30 7
40 4
50 4
60 4
70 4
80 2
90 0
100 0
};
\addlegendentry{GATO-CT}
\end{axis}

\end{tikzpicture}}
        \label{sfig:meta45_sr}
    \end{subfigure}
    \vspace{-1.em}
    \caption{Comparisons of \textbf{generalist agents} on \textit{MetaWorld 45 tasks} on Percentage of Expert Score (PES) (left) and Success Rate (SR) (right).}
    \label{fig:g-meta45}
    \vspace{-.5em}
\end{figure}
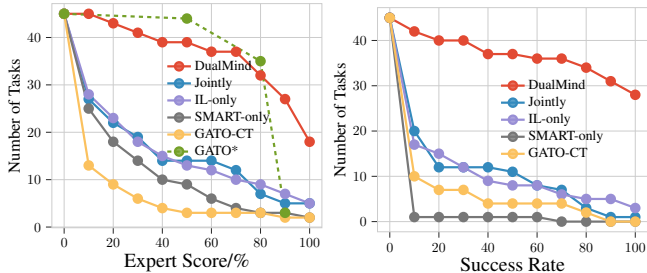

In this section, we aim to demonstrate the capabilities of \texttt{\ours} on all tasks. Note that, as a generalist agent, the performance on both MetaWorld and Habitat are achieved by a single model. 
The performance is shown in Fig.~\ref{fig:g-hab} and Fig.~\ref{fig:g-meta45}. To provide a reference for readers, we follow GATO's evaluation protocol and report the Percentage Expert Score (PES), which measures the number of distinct tasks for which each model performs above a given score threshold relative to the expert performance. For each task, we roll out the model 10 times and average the defined scores. As shown in Fig.~\ref{fig:g-meta45}, \texttt{\ours} achieves over 90$\%$ expert score threshold across more than 27 tasks, outperforming \texttt{GATO*} by a large margin, which only has three tasks above the threshold. On lower expert score thresholds, for example, 80$\%$ and 50$\%$, \texttt{\ours} can also achieve comparable performance. However, it should be noted that GATO's performance was achieved by their 1.18B model trained on massive datasets. Therefore, this is just a reference for readers, and a fully fair comparison with GATO cannot be performed without access to both the model and data. To provide a more fair comparison, we compare \texttt{\ours} with a self-implemented GATO (\texttt{GATO-CT}), which will be discussed in more detail in Sec.~\ref{ssec:comp-baseline}. We also report the number of tasks for which our model performs above a given Success Rate (SR). \texttt{\ours} achieves 39 tasks at over 0.5 SR and can maintain good performance on higher SRs, with 34 tasks at over 0.8 SR and 28 tasks at over 1 SR.
We present the performance of \texttt{\ours} on Habitat by averaging across all 12 testing scenes and reporting the success rate (SR) and success weighted by path length (SPL)~\cite{anderson2018evaluation} evaluation metrics. As shown Fig.~\ref{fig:g-hab}, \texttt{\ours} outperforms the other baseline models by a large margin under both evaluation metrics. (See performance on each task in Appendix~\ref{app:more-exp}.)


\begin{figure}[t!]
    \centering
    \begin{subfigure}[t]{0.47\columnwidth}
        \centering
        \resizebox{1.1\columnwidth}{!}{
\begin{tikzpicture}

\begin{axis}[
width=10cm,
height=8cm,
axis line style={white!80!black},
tick align=outside,
tick pos=left,
x grid style={white!80!black},
xmin=0.0725, xmax=0.65,
xtick style={color=white!15!black},
xtick={0.15,0.3,0.45,0.6},
xticklabel style={rotate=15, font=\fontsize{14}{6}\selectfont},
xticklabels={\textbf{\ours/single},\textbf{Jointly/single},\textbf{IL-only/single},\textbf{SMART-only/single},},
y grid style={white!80!black},
ylabel={\LARGE{Success Rate}},
ylabel style={yshift=6pt},
ymajorgrids,
ymin=0, ymax=0.189,
ytick style={color=white!15!black},
ytick={0.0,0.05,0.1,0.15},
yticklabel style={ font=\fontsize{14}{6}\selectfont},
yticklabels={0.0,0.05,0.1,0.15},
]
\draw[draw=none,fill=ours] (axis cs:0.1,0) rectangle (axis cs:0.2,0.148);
\draw[draw=none,fill=jointly] (axis cs:0.25,0) rectangle (axis cs:0.35,0.18);
\draw[draw=none,fill=il] (axis cs:0.4,0) rectangle (axis cs:0.5,0.057);
\draw[draw=none,fill=white!41.1764705882353!black] (axis cs:0.55,0) rectangle (axis cs:0.65,0);
\draw (axis cs:0.575,0.001) node[
  scale=2,
  anchor=base west,
  text=black,
  rotate=0.0
]{\bfseries 0};
\end{axis}

\end{tikzpicture}} 
        \label{sfig:hat_sr_single}
    \end{subfigure}
    \hfill
    \begin{subfigure}[t]{0.47\columnwidth}
        \centering
        \resizebox{1.1\columnwidth}{!}{
\begin{tikzpicture}

\begin{axis}[
width=10cm,
height=8cm,
axis line style={white!80!black},
tick align=outside,
tick pos=left,
x grid style={white!80!black},
xmin=0.0725, xmax=0.65,
xtick style={color=white!15!black},
xtick={0.15,0.3,0.45,0.6},
xticklabel style={rotate=15, font=\fontsize{14}{6}\selectfont},
xticklabels={\textbf{\ours/single},\textbf{Jointly/single},\textbf{IL-only/single},\textbf{SMART-only/single},},
y grid style={white!80!black},
ylabel={\LARGE{SPL}},
ylabel style={yshift=6pt},
ymajorgrids,
ymin=0, ymax=0.147,
ytick style={color=white!15!black},
yticklabel style={ font=\fontsize{14}{6}\selectfont},
ytick={0.0,0.05,0.1,0.15},
yticklabels={0.0,0.05,0.1,0.15},
]
\draw[draw=none,fill=ours] (axis cs:0.1,0) rectangle (axis cs:0.2,0.136);
\draw[draw=none,fill=jointly] (axis cs:0.25,0) rectangle (axis cs:0.35,0.14);
\draw[draw=none,fill=il] (axis cs:0.4,0) rectangle (axis cs:0.5,0.043);
\draw[draw=none,fill=white!41.1764705882353!black] (axis cs:0.55,0) rectangle (axis cs:0.65,0);
\draw (axis cs:0.575,0.001) node[
  scale=2,
  anchor=base west,
  text=black,
  rotate=0.0
]{\bfseries 0};
\end{axis}

\end{tikzpicture}}
        \label{sfig:hat_spl_single}
    \end{subfigure}
    \vspace{-2em}
    \caption{Comparisons of \textbf{single-domain specialist} on \textit{Habitat 12 scenes}, as measured by  SR (left) and SPL (right).}
    \label{fig:g-hab-single}
\end{figure}
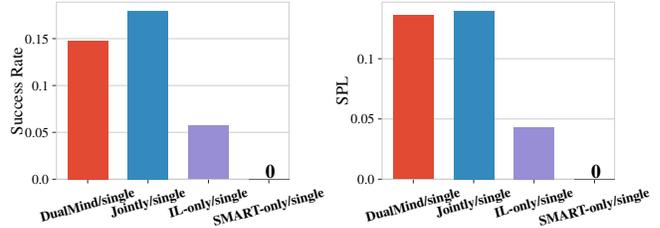


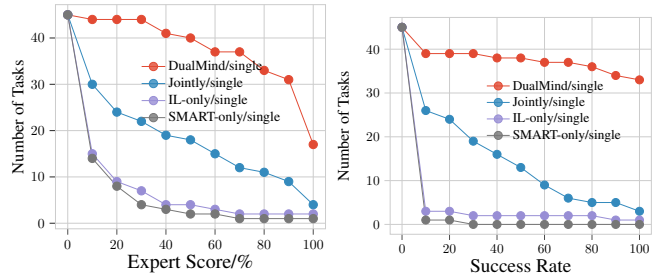
\begin{figure}[t!]
    \centering
    \begin{subfigure}[t]{0.47\columnwidth}
        \centering
        \resizebox{1.1\columnwidth}{!}{
\begin{tikzpicture}

\begin{axis}[
axis line style={white!80!black},
legend cell align={left},
legend style={fill opacity=0.8, draw opacity=1, text opacity=1, at={(0.58,0.61)}, anchor=center, draw=none, fill=none},
tick align=outside,
tick pos=left,
x grid style={white!80!black},
xlabel={\Large{Expert Score/\%}},
xmajorgrids,
xmin=-5, xmax=105,
xtick style={color=white!30!black},
y grid style={white!80!black},
y label style={at={(axis description cs:0.08,0.5)},anchor=south},
ylabel={\large{Number of Tasks}},
ymajorgrids,
ymin=-1.2, ymax=47.2,
ytick style={color=white!15!black}
]
\addplot [semithick, ours, mark=*, mark size=3, mark options={solid}]
table {%
0 45
10 44
20 44
30 44
40 41
50 40
60 37
70 37
80 33
90 31
100 17
};
\addlegendentry{\ours/single}
\addplot [semithick, jointly, mark=*, mark size=3, mark options={solid}]
table {%
0 45
10 30
20 24
30 22
40 19
50 18
60 15
70 12
80 11
90 9
100 4
};
\addlegendentry{Jointly/single}
\addplot [semithick, il, mark=*, mark size=3, mark options={solid}]
table {%
0 45
10 15
20 9
30 7
40 4
50 4
60 3
70 2
80 2
90 2
100 2
};
\addlegendentry{IL-only/single}
\addplot [semithick, white!46.6666666666667!black, mark=*, mark size=3, mark options={solid}]
table {%
0 45
10 14
20 8
30 4
40 3
50 2
60 2
70 1
80 1
90 1
100 1
};
\addlegendentry{SMART-only/single}
\end{axis}

\end{tikzpicture}} 
        \label{sfig:meta_es_single}
    \end{subfigure}
    \hfill
    \begin{subfigure}[t]{0.47\columnwidth}
        \centering
        \resizebox{1.1\columnwidth}{!}{
\begin{tikzpicture}

\begin{axis}[
axis line style={white!80!black},
legend cell align={left},
legend style={fill opacity=0.8, draw opacity=1, text opacity=1, at={(0.91,0.57)}, anchor=east, draw=none, fill=none},
tick align=outside,
tick pos=left,
x grid style={white!80!black},
xlabel={\Large{Success Rate}},
xmajorgrids,
xmin=-5, xmax=105,
xtick style={color=white!30!black},
y grid style={white!80!black},
ylabel={\large{Number of Tasks}},
ymajorgrids,
ymin=-2.25, ymax=47.25,
ytick style={color=white!80!black}
]
\addplot [semithick, ours, mark=*, mark size=3, mark options={solid}]
table {%
0 45
10 39
20 39
30 39
40 38
50 38
60 37
70 37
80 36
90 34
100 33
};
\addlegendentry{\ours/single}
\addplot [semithick, jointly, mark=*, mark size=3, mark options={solid}]
table {%
0 45
10 26
20 24
30 19
40 16
50 13
60 9
70 6
80 5
90 5
100 3
};
\addlegendentry{Jointly/single}
\addplot [semithick, il, mark=*, mark size=3, mark options={solid}]
table {%
0 45
10 3
20 3
30 2
40 2
50 2
60 2
70 2
80 2
90 1
100 1
};
\addlegendentry{IL-only/single}
\addplot [semithick, white!46.6666666666667!black, mark=*, mark size=3, mark options={solid}]
table {%
0 45
10 1
20 1
30 0
40 0
50 0
60 0
70 0
80 0
90 0
100 0
};
\addlegendentry{SMART-only/single}
\end{axis}

\end{tikzpicture}}
        \label{sfig:meta_sr_single}
    \end{subfigure}
    \vspace{-1.em}
    \caption{Comparisons of \textbf{single-domain specialist} on \textit{MetaWorld 45 tasks}, as measured by Percentage of Expert Score (PES) (left) and Success Rate (SR) (right).}
    \label{fig:g-meta-single}
    \vspace{-.5em}
\end{figure}

\subsection{Analysis}

\subsubsection{Different training regimes} \label{ssec:train-appr}

\begin{figure*}[htbp!]
    \centering
    \includegraphics[width=1.\textwidth]{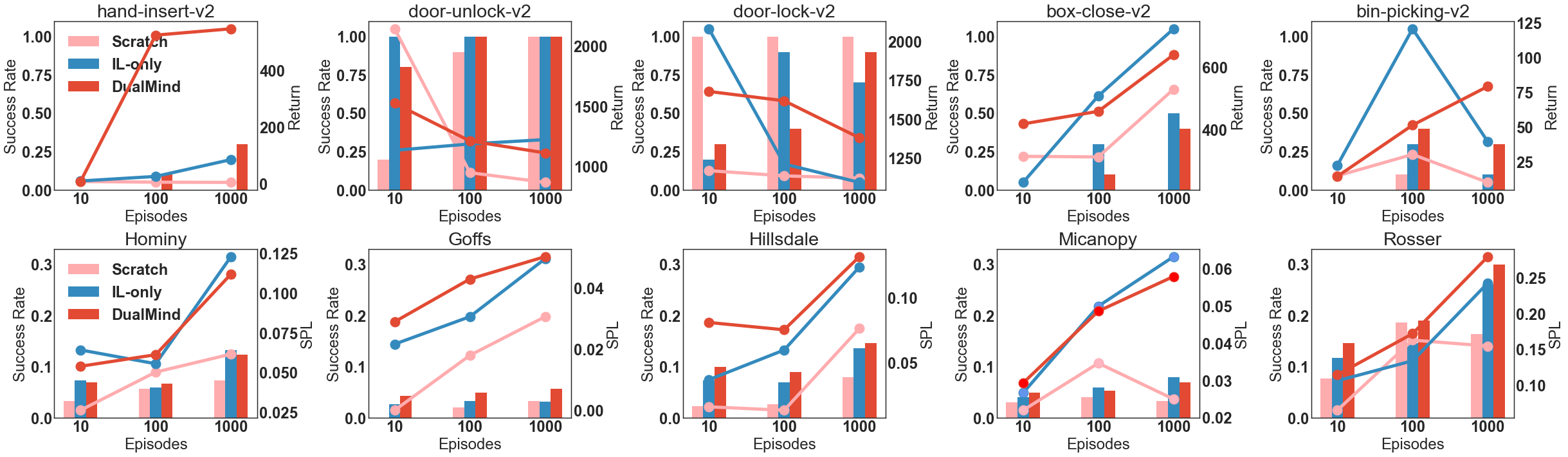}
    \vspace{-2em}
    \caption{Few-shot comparisons of generalist agents on out-of-distribution tasks. The performance of success rate (left axis, bar-chart) or Return/SPL (right axis, line-chart) on different tasks after we performed 10-, 100-, and 1000-shot  learning on \texttt{\ours}(\textcolor{ours}{red}), \texttt{IL-only}(\textcolor{blue}{blue}), and \texttt{Scratch}(\textcolor{spink}{pink}). Specifically, the bar charts (on the left axis) represent success rates, while the line charts (on the right axis) represent SPL in Habitat or Return in Metaworld.}
    \label{fig:g-ood}
\end{figure*}

\subsubsection*{Is imitation learning all you need for a generalist agent?} 
To answer the question, in this experiment, we compare \texttt{\ours} with its counterpart trained only with Imitation Learning objective, i.e., \texttt{IL-only}. In Fig.~\ref{fig:g-hab} and Fig.~\ref{fig:g-meta45}, we present the comparison results between the generalist multi-domain agents. As shown in the figures, \texttt{\ours} outperforms its \texttt{IL-only} counterpart by over 50$\%$ and 70$\%$ on Habitat and MetaWorld, respectively. Specifically, \texttt{\ours} performs well on 39 out of 45 tasks over the 50$\%$ expert score threshold, while \texttt{IL-only} only performs well on 13 tasks. As the difficulty of the tasks increases, \ours still maintains good performance, achieving 18 tasks and 28 tasks at the 100$\%$ expert score and SR, respectively, while \texttt{IL-only} only achieves 5 tasks. Similar observations can also be made when comparing the single-domain specialist agents (Fig.~\ref{fig:g-meta-single} and Fig.~\ref{fig:g-hab-single}). (See performance on each tasks in Appendix~\ref{app:more-exp}.)

We can infer from this that Imitation Learning alone may not suffice to build a truly general-purpose model, particularly when aiming to tackle tasks that span a broad range of domains. Even within a single domain, variations in embodiments, scenes, and instructions can pose significant challenges. We conducted additional investigations into the generalization capabilities by comparing different approaches on out-of-distribution tasks, as demonstrated in Section~\ref{ssec:ablate-ood}. 


\subsubsection*{Can self-supervised learning well-align with instructions without FT?}

To address this inquiry, we compare \texttt{\ours} with its self-supervised equivalent, \texttt{SMART-only}, while also evaluating both single- and multi-domain agents. As depicted in Fig.~\ref{fig:g-hab} and Fig.~\ref{fig:g-meta45}, \texttt{\ours} exhibits superior performance compared to \texttt{SMART-only}, with over 75$\%$ and 78$\%$ better results on Habitat and MetaWorld, respectively. Notably, \texttt{SMART-only} is unable to succeed on any tasks when applied to single-domain agents, whereas \texttt{\ours} maintains a significant advantage, particularly on MetaWorld.

Our hypothesis is that SMART, being a pretrain-finetune pipeline, is unlikely to attain the desired performance without post-finetuning. Even when training \texttt{SMART-only} by providing prompts in the same manner as \texttt{\ours}, zero-shot prompting may not be achievable due to limitations in the self-supervised training objective not being well-aligned with task instructions, as detailed in Section~\ref{sec:prelim}. Additionally, we noted that \texttt{SMART-only} surpasses its single-domain equivalent, suggesting its effectiveness in capturing shared knowledge across diverse data.

\subsubsection*{Do we need to train them in two phases?}

As \ours is trained using different objectives in two phases, one may question the necessity of such an approach. Firstly, from an optimization standpoint, training all four losses jointly may present more challenges in terms of steady optimization. Different optimization directions could potentially conflict with each other, and varying convergence rates could hinder all objectives from being trained to reach optimality.
Furthermore, in terms of computational costs, \texttt{\ours} only needs to optimize a small portion of the model weights in phase 2 (as demonstrated in the ablations presented in Section~\ref{ssec:ablate-phase}). This makes the training process more efficient and cost-effective compared to its jointly trained counterpart. In this experiment, we provide further empirical evidence to support this claim.

As illustrated in Fig.~\ref{fig:g-hab}, Fig.~\ref{fig:g-meta45}, Fig.~\ref{fig:g-hab-single}, and Fig.~\ref{fig:g-meta-single}, \texttt{Jointly} outperforms \texttt{IL-only} and \texttt{SMART-only}, thereby confirming the necessity of utilizing all training objectives. However, it lags behind \ours by a considerable margin in both multi- and single-domain comparisons.
Interestingly, \texttt{Jointly} slightly outperforms \texttt{\ours} in single-domain comparisons. We hypothesize that the optimization challenges may not be as significant as those encountered when training on data from the same domain.

\subsubsection{Out-of-distribution tasks}{\label{ssec:ablate-ood}}

The objective of this experiment is to assess the ability of our model to solve novel tasks. To achieve this, we evaluate our models on 10 held-out tasks from two domains, namely MetaWorld and Habitat. 
The MetaWorld tasks consist of ``hand-insert-v2", ``door-unlock-v2", ``door-lock-v2", ``box-close-v2", and ``bin-picking-v2", whereas the Habitat tasks include ``Goffs", ``Hominy", ``Hillsdale", ``Micanopy", and ``Rosser". 
To evaluate the performance of our models, we follow the evaluation protocol with GATO, which involves finetuning each agent on a limited number of demonstrations. Specifically, we conduct 10-, 100-, and 1000-shot learning and finetune all models for 10000
gradient steps. Further details on the evaluation protocol can be found in Appendix~\ref{app:imp-detail}.


We compare the performance of three models, namely \texttt{\ours}, \texttt{IL-only}, and \texttt{Scratch}. \texttt{\ours} and \texttt{IL-only} are finetuned with few-shot demonstrations from the base model of \ours and \texttt{IL-only}, respectively. In contrast, \texttt{Scratch} refers to the model that is trained on few-shot demonstrations from randomly initialized model weights. Fig.~\ref{fig:g-ood} illustrates the success rate (shown on the left axis using a bar chart) and Return/SPL (shown on the right axis using a line chart) across different tasks after implementing 10-shot, 100-shot, and 1000-shot learning on these models. 
As demonstrated in Fig.~\ref{fig:g-ood}, \texttt{Scratch} performs the worst among the three models in most cases.
Upon comparing \ours with \texttt{IL-only}, we observe that \ours exhibits superior performance across various shot settings. Specifically, in terms of the SR metric, \ours outperforms \texttt{IL-only} on 8 out of 10 tasks at 10-shot and on 7 tasks at 100- and 1000-shot demonstrations. Furthermore, with respect to the SPL and PES metrics, \ours achieves better results than \texttt{IL-only} on 9 tasks in the 10-shot experiment. These results provide further evidence that the proposed \duelph training approach can enhance the generalization ability of models even when dealing with novel tasks and limited demonstrations.

\begin{figure}[t!]
    \centering
    \includegraphics[width=.48\textwidth]{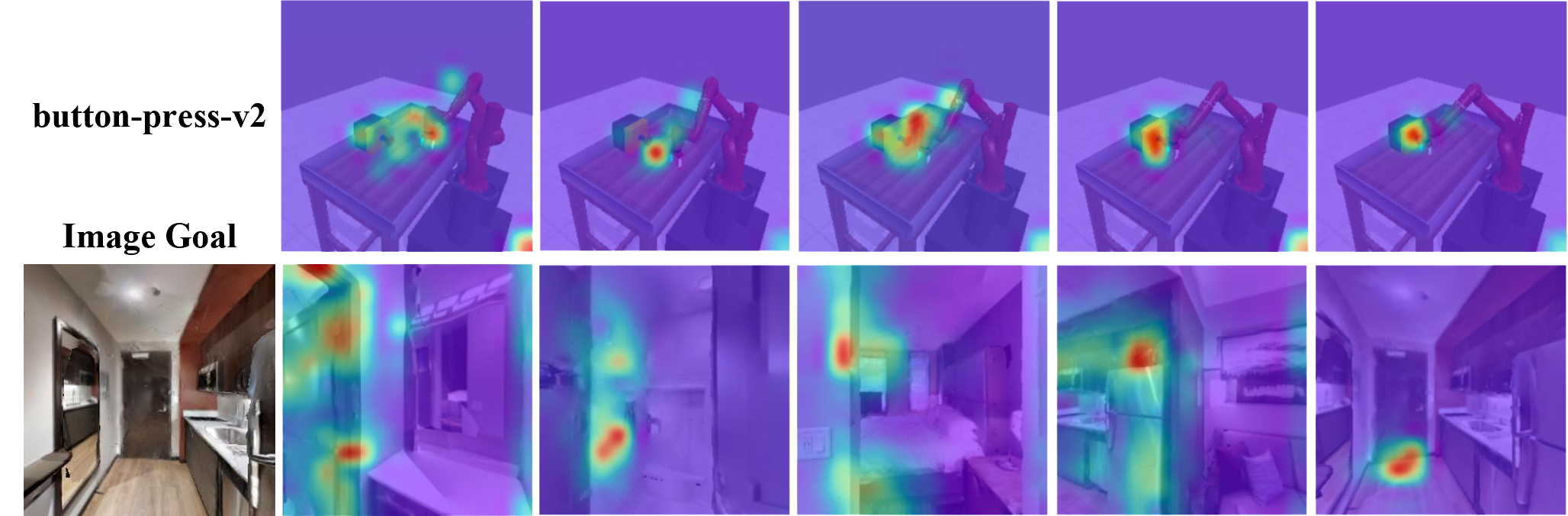}
    \caption{Attention map visualization.}
    \label{fig:atten_vis}
    \vspace{-1.5em}
\end{figure}
\subsubsection{Attention visualization} \label{ssec:attention}

To gain insight into how \texttt{\ours} is able to perform diverse tasks, we conduct attention visualization. We present attention maps for tasks from both Habitat and MetaWorld, where we display a sequence of frames from the episode for each task.

The attention maps reveal that when performing manipulation tasks in MetaWorld, such as and ``button-press-v2", the model initially focuses on the execution context and then shifts its attention to the targeting instance, such as the ``button", until the task is completed. Notably, for navigation tasks in Habitat, \texttt{\ours} learns to explore the scene to locate the goal. For example, as shown in Fig.~\ref{fig:atten_vis}, given an image goal, the agent first attends to the entrance to navigate into the restroom. Upon realizing that the goal is not there, it steps out and searches for another room to enter. After spotting the refrigerator, which appears in the image goal, the agent quickly locks onto the goal and completes the task. These attention maps provide insight into how \texttt{\ours} leverages its generalization ability to solve new tasks.

\subsection{Ablation study} {\label{ssec:ablate}}

\subsubsection*{Training parts in Phase \Romannum{2}}
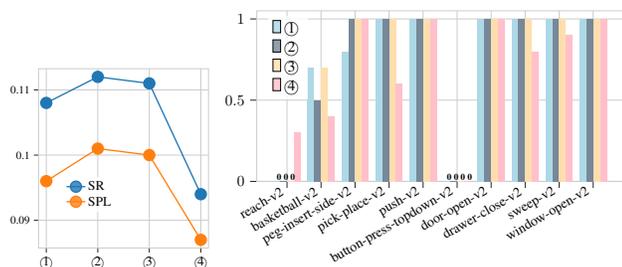
\begin{figure}[t!]
    \centering
    \begin{subfigure}[t]{0.33\columnwidth}
        \centering
        \resizebox{\columnwidth}{!}{
\begin{tikzpicture}

\definecolor{color0}{rgb}{0.12156862745098,0.466666666666667,0.705882352941177}
\definecolor{color1}{rgb}{1,0.498039215686275,0.0549019607843137}

\begin{axis}[
width=7cm,
height=7.3cm,
axis line style={white!80!black},
legend cell align={left},
legend style={fill opacity=0.8, draw opacity=1, text opacity=1, at={(0.3,0.3)}, anchor=center, draw=none, font=\large},
tick align=outside,
tick pos=left,
x grid style={white!80!black},
xmajorgrids,
xmin=0.85, xmax=4.15,
xtick style={color=white!15!black},
xtick={1,2,3,4},
xticklabels={\large{\textcircled{1}},\large{\textcircled{2}},\large{\textcircled{3}},\large{\textcircled{4}}},
y grid style={white!80!black},
ymajorgrids,
ymin=0.08575, ymax=0.11325,
ytick style={color=white!15!black},
ytick={0.09,0.1,0.11},
yticklabels={0.09,0.1,0.11}
]
\addplot [very thick, color0, mark=*, mark size=5, mark options={solid}]
table {%
1 0.108
2 0.112
3 0.111
4 0.094
};
\addlegendentry{SR}
\addplot [very thick, color1, mark=*, mark size=5, mark options={solid}]
table {%
1 0.096
2 0.101
3 0.1
4 0.087
};
\addlegendentry{SPL}
\end{axis}

\end{tikzpicture}} 
        \label{sfig:freeze_line}
    \end{subfigure}
    \hfill
    \begin{subfigure}[t]{0.65\columnwidth}
        \centering
        \resizebox{\columnwidth}{!}{
\begin{tikzpicture}

\pgfplotsset{compat=1.11,
    /pgfplots/ybar legend/.style={
    /pgfplots/legend image code/.code={%
       \draw[##1,/tikz/.cd,yshift=-0.25em]
        (0cm,0cm) rectangle (5pt,0.8em);},
   },
}

\definecolor{color1}{rgb}{0.466666666666667,0.533333333333333,0.6}
\definecolor{color0}{rgb}{0.67843137254902,0.847058823529412,0.901960784313726}
\definecolor{color2}{rgb}{1,0.870588235294118,0.67843137254902}
\definecolor{color3}{rgb}{1,0.752941176470588,0.796078431372549}

\begin{axis}[
width=10cm,
height=5.5cm,
axis line style={white!80!black},
legend cell align={left},
legend style={
  fill opacity=0.8,
  draw opacity=1,
  text opacity=1,
  at={(0.03,0.97)},
  anchor=north west,
  draw=none
},
tick align=outside,
tick pos=left,
x grid style={white!80!black},
xmajorgrids,
xmin=-0.89, xmax=9.89,
xtick style={color=white!15!black},
xtick={0,1,2,3,4,5,6,7,8,9},
xticklabel style={rotate=30.0,anchor=east},
xticklabels={
  \small{reach-v2},
  \small{basketball-v2},
  \small{peg-insert-side-v2},
  \small{pick-place-v2},
  \small{push-v2},
  \small{button-press-topdown-v2},
  \small{door-open-v2},
  \small{drawer-close-v2},
  \small{sweep-v2},
  \small{window-open-v2}
},
y grid style={white!80!black},
ymajorgrids,
ymin=0, ymax=1.05,
ytick style={color=white!15!black}
]
\draw[draw=none,fill=color0] (axis cs:-0.4,0) rectangle (axis cs:-0.2,0);
\addlegendimage{ybar,ybar legend,draw=none,fill=color0}
\addlegendentry{\textcircled{1}}

\draw[draw=none,fill=color0] (axis cs:0.6,0) rectangle (axis cs:0.8,0.7);
\draw[draw=none,fill=color0] (axis cs:1.6,0) rectangle (axis cs:1.8,0.8);
\draw[draw=none,fill=color0] (axis cs:2.6,0) rectangle (axis cs:2.8,1);
\draw[draw=none,fill=color0] (axis cs:3.6,0) rectangle (axis cs:3.8,1);
\draw[draw=none,fill=color0] (axis cs:4.6,0) rectangle (axis cs:4.8,0);
\draw[draw=none,fill=color0] (axis cs:5.6,0) rectangle (axis cs:5.8,1);
\draw[draw=none,fill=color0] (axis cs:6.6,0) rectangle (axis cs:6.8,1);
\draw[draw=none,fill=color0] (axis cs:7.6,0) rectangle (axis cs:7.8,1);
\draw[draw=none,fill=color0] (axis cs:8.6,0) rectangle (axis cs:8.8,1);
\draw[draw=none,fill=color1] (axis cs:-0.2,0) rectangle (axis cs:0,0);
\addlegendimage{ybar,ybar legend,draw=none,fill=color1}
\addlegendentry{\textcircled{2}}

\draw[draw=none,fill=color1] (axis cs:0.8,0) rectangle (axis cs:1,0.5);
\draw[draw=none,fill=color1] (axis cs:1.8,0) rectangle (axis cs:2,1);
\draw[draw=none,fill=color1] (axis cs:2.8,0) rectangle (axis cs:3,1);
\draw[draw=none,fill=color1] (axis cs:3.8,0) rectangle (axis cs:4,1);
\draw[draw=none,fill=color1] (axis cs:4.8,0) rectangle (axis cs:5,0);
\draw[draw=none,fill=color1] (axis cs:5.8,0) rectangle (axis cs:6,1);
\draw[draw=none,fill=color1] (axis cs:6.8,0) rectangle (axis cs:7,1);
\draw[draw=none,fill=color1] (axis cs:7.8,0) rectangle (axis cs:8,1);
\draw[draw=none,fill=color1] (axis cs:8.8,0) rectangle (axis cs:9,1);
\draw[draw=none,fill=color2] (axis cs:-1.38777878078145e-17,0) rectangle (axis cs:0.2,0);
\addlegendimage{ybar,ybar legend,draw=none,fill=color2}
\addlegendentry{\textcircled{3}}

\draw[draw=none,fill=color2] (axis cs:1,0) rectangle (axis cs:1.2,0.7);
\draw[draw=none,fill=color2] (axis cs:2,0) rectangle (axis cs:2.2,1);
\draw[draw=none,fill=color2] (axis cs:3,0) rectangle (axis cs:3.2,1);
\draw[draw=none,fill=color2] (axis cs:4,0) rectangle (axis cs:4.2,1);
\draw[draw=none,fill=color2] (axis cs:5,0) rectangle (axis cs:5.2,0);
\draw[draw=none,fill=color2] (axis cs:6,0) rectangle (axis cs:6.2,1);
\draw[draw=none,fill=color2] (axis cs:7,0) rectangle (axis cs:7.2,1);
\draw[draw=none,fill=color2] (axis cs:8,0) rectangle (axis cs:8.2,1);
\draw[draw=none,fill=color2] (axis cs:9,0) rectangle (axis cs:9.2,1);
\draw[draw=none,fill=color3] (axis cs:0.2,0) rectangle (axis cs:0.4,0.3);
\addlegendimage{ybar,ybar legend,draw=none,fill=color3}
\addlegendentry{\textcircled{4}}

\draw[draw=none,fill=color3] (axis cs:1.2,0) rectangle (axis cs:1.4,0.4);
\draw[draw=none,fill=color3] (axis cs:2.2,0) rectangle (axis cs:2.4,1);
\draw[draw=none,fill=color3] (axis cs:3.2,0) rectangle (axis cs:3.4,0.6);
\draw[draw=none,fill=color3] (axis cs:4.2,0) rectangle (axis cs:4.4,1);
\draw[draw=none,fill=color3] (axis cs:5.2,0) rectangle (axis cs:5.4,0);
\draw[draw=none,fill=color3] (axis cs:6.2,0) rectangle (axis cs:6.4,1);
\draw[draw=none,fill=color3] (axis cs:7.2,0) rectangle (axis cs:7.4,0.8);
\draw[draw=none,fill=color3] (axis cs:8.2,0) rectangle (axis cs:8.4,0.9);
\draw[draw=none,fill=color3] (axis cs:9.2,0) rectangle (axis cs:9.4,1);
\draw (axis cs:-0.39,0.01) node[
  scale=0.6,
  anchor=base west,
  text=black,
  rotate=0.0
]{\bfseries 0};
\draw (axis cs:4.61,0.01) node[
  scale=0.6,
  anchor=base west,
  text=black,
  rotate=0.0
]{\bfseries 0};
\draw (axis cs:-0.19,0.01) node[
  scale=0.6,
  anchor=base west,
  text=black,
  rotate=0.0
]{\bfseries 0};
\draw (axis cs:4.81,0.01) node[
  scale=0.6,
  anchor=base west,
  text=black,
  rotate=0.0
]{\bfseries 0};
\draw (axis cs:0.01,0.01) node[
  scale=0.6,
  anchor=base west,
  text=black,
  rotate=0.0
]{\bfseries 0};
\draw (axis cs:5.01,0.01) node[
  scale=0.6,
  anchor=base west,
  text=black,
  rotate=0.0
]{\bfseries 0};
\draw (axis cs:5.21,0.01) node[
  scale=0.6,
  anchor=base west,
  text=black,
  rotate=0.0
]{\bfseries 0};
\end{axis}

\end{tikzpicture}}
        \label{sfig:freeze_task}
    \end{subfigure}
    \vspace{-2em}
    \caption{Comparisons of frozen parts in Phase \Romannum{2}. }
    \label{fig:freeze}
    \vspace{-1em}
\end{figure}

{\label{ssec:ablate-phase}}
In this section, we ablate \ours by varying model weights that been trained in Phase \Romannum{2}, as listed below: 
\begin{itemize}[noitemsep,leftmargin=*,topsep=0pt]
    \item \textcircled{1}: Freeze the entire \ourmod arch by only train the cross-attention layers.
    \item \textcircled{2}: Freeze the Transformer Encoder (State tokenizer) and the first 4 layers of Transformer Decoder. 
    \item \textcircled{3}: Freeze the Transformer Encoder.
    \item \textcircled{4}: No frozen part, optimize the entire model in Phase \Romannum{2}.
\end{itemize}
As shown in Fig.~\ref{fig:freeze}, \textcircled{2} and \textcircled{3} perform the best in most cases. For our experiments, we use \textcircled{3}. However, for future scaled-up models and data, we would recommend using \textcircled{2} since it saves more computational cost. When training each setting with the same number of iterations, \textcircled{4} performs poorly, which may be due to slow convergence with more model weights. This result also suggests that after training in phase \Romannum{1}, our model has learned useful information, but insufficient re-training in phase \Romannum{2} may lead to performance deterioration due to potential forgetting issues.

\subsubsection*{Prompt conditioning}{\label{ssec:ablate-prompt}}

We conducted an ablation study on \ours by comparing two prompt conditioning approaches: prefix and \xatten prompting. We used the average success rate of ML10 training tasks as the comparison metric for Metaworld. Results show that \xatten prompting achieves a 0.76 SR on Metaworld and an 0.11 SR on Habitat, while prefix prompting only achieves 0.29 SR and 0 SR, respectively. The cross-attention mechanism in \xatten prompting allows the agent to establish a strong connection between prompts and demonstrations, which is particularly useful for goal-conditioned tasks. (See more details and discussion in Appendix~\ref{app:more-exp}.)

\section{Conclusion}

This paper presents a new training approach for generalist agents called \ours, which consists of two phases: self-supervised learning of basic and generic knowledge across various tasks, followed by imitation of expert behaviors with different types of prompt conditioning.
By utilizing a carefully designed Transformer Encoder-Decoder architecture and a dual-phase training scheme, \ours is scalable, versatile, and generalizable.
Empirical evaluation on two challenging domains, Habitat and MetaWorld, shows that \ours outperforms previous generalist learning methods and pretraining approaches.
Further analysis and ablations demonstrate the effectiveness of the dual-phase design.

Future work includes expanding \ours to more domains and tasks, finding efficient solutions for handling longer context lengths in demonstrations, and enabling practical training in online interactive scenarios.

{\small
\bibliographystyle{ieee_fullname}
\bibliography{egbib}
}

\clearpage
\appendix
\onecolumn
{\centering{\Large Appendix}}
\section*{Model Card of \ours}

\begin{longtable}{p{4cm}|p{12cm}}
    \toprule
        \multicolumn{2}{c}{\textbf{Model Details}} \\
    \hline
             Model Type    &  Encoder-Decoder Transformer (\ourmod), built upon a ViT encoder~\cite{dosovitskiy2020image}, a TokenLearner~\cite{NEURIPS2021_6a30e32e:tokenlearner}, and a Control Transformer~\cite{sun2023smart}. \\\hline
             Training Process    & The training process is divided into two phases. In Phase \Romannum{1}, the entire \ourmod is trained with a self-supervised training objective. In Phase \Romannum{2},a small part of the model is trained using imitation learning with prompt conditions. The detailed training objective is described in Sec.~\ref{ssec:obj}.\\\hline
             Model Version & Initial release. \\ 
    \midrule
    \multicolumn{2}{c}{\textbf{Intended Uses}} \\
    \hline
            Primary Intended Uses & The proposed model aims to perform a wide range of control tasks spanning multiple domains, visual scenes and embodiments. Our intention is to create a general-purpose decision-making solution capable of handling various tasks using a single set of weights, without requiring task-specific fine-tuning.  \\
    \midrule
    \multicolumn{2}{c}{\textbf{Factors}} \\
    \hline
            Relevant Factors & Multiple factors can influence the performance of the model. First, the quality of training dataset has influence on the results, including task diversity, behavior policy performance, data volume, etc. Second, model implementation hyperparameter setting, and training objectives will also alter the final performance.  \\\hline
            Evaluation Factors & We report the performance of the model in multiple sets of tasks, and conducted ablation study in Sec.~\ref{ssec:ablate-ood}. \\
    \midrule
    \multicolumn{2}{c}{\textbf{Metrics}} \\
    \hline
            Model Performance Measures & Our downstream task performance is measured using success rate, SPL, and expert score, as detailed in Sec.~\ref{supp:eval_detail}. The expert score is calculated in the same manner as GATO~\cite{reed2022generalist}, while using a different dataset. \\\hline
            Decision thresholds  & N/A\\\hline
            Approaches to Uncertainty and Variability & The model evaluation process inevitably involves uncertainties. In order to reduce the variance introduced during the evaluation, we employed 3 random seeds for the Habitat evaluation and 10 random seeds for the Metaworld evaluation. \\
    \midrule
    \multicolumn{2}{c}{\textbf{Evaluation Data}} \\
    \hline
            Datasets & Our \ours is evaluated on multiple control tasks from Habitat and Metaworld. Both in-distribution and out-of-distribution tasks are considered. 
            \textbf{Habitat:} Our experiments are focused on the ImageNav task, we chose 4 scenes . We hold out 5 Gibson scenes for the experiments of out-of-distribution tasks, as detailed in Sec.~\ref{supp:eval_detail}.
            \textbf{Metaworld:}  We select 45 training tasks on ML45 for the experiments of evaluation, and hold out 5 test tasks for the experiments of out-of-distribution, as detailed in Sec.~\ref{supp:eval_detail}. \\\hline
            Motivation & Our evaluation of \ours consists of two components. First, we evaluated its performance on in-distribution tasks to understand how well it handles tasks across domains, scenes, and embodiments using a single set of model weights. Second, we evaluated \ours on out-of-distribution tasks to assess its ability to adapt to entirely new tasks. \\\hline
            Preprocessing & Observations are tokenized into the same embedding sequence before being input to transformer decoder, as detailed in Sec.~\ref{ssec:model}.\\
    \midrule
    \multicolumn{2}{c}{\textbf{Training Data}} \\
    \hline
            Datasets & The model is trained using 100K episodes collected from Habitat and Metaworld, with 50k episodes ($\sim$3.26M interaction steps) on Habitat and 50K episodes ($\sim$3.82M interaction steps) on Metaworld, respectively. 
            \\\hline
            Motivation & In order to ensure that \ours can handle tasks across domains, scenes, and embodiments, we collected data for all tasks in Metaworld and all scenes in Habitat. The data collection process is detailed in Sec.~\ref{sup:data_collect}.
  \\\hline
            Preprocessing & The multi-domain data is tokenized into the same embedding sequence before being fed to the transformer decoder, as detailed in Sec.~\ref{ssec:model}. \\
    \midrule
    \multicolumn{2}{c}{\textbf{Quantitative Analyses}} \\
    \hline
            Unitary Results &  We evaluated the performance of \ours on the Metaworld and Habitat benchmarks. In Sections \ref{sec:capabilities} and \ref{ssec:train-appr}, we demonstrate the general capabilities of \ours across both Metaworld and Habitat tasks. Additionally, in Section \ref{ssec:ablate-ood}, we analyze its performance on out-of-distribution tasks. \\ 
    \midrule
    \multicolumn{2}{c}{\textbf{Ethical Considerations}} \\
    \hline
            Data & Our data is collected from simulators of navigation and manipulation, and thus it does not include any unethical data.\\\hline
            Risks and Harms & Our current training and evaluation are conducted in simulators, and do not involve physical robots where model malfunctioning could lead to safety issues.  \\\hline
            Mitigations & N/A \\
    \midrule
    \multicolumn{2}{c}{\textbf{Caveats and Recommendation}} \\
    \hline 
            Future work & Our future work includes expanding \ours to more domains and tasks, finding efficient solutions for handling longer context lengths in demonstrations, and enabling practical training in online interactive scenarios. \\
    \bottomrule
\caption{Model card of \ours, following the framework proposed by~\cite{mitchell2019model}.}
\label{tab:model_card}
\end{longtable}

\section{Implementation details}{\label{app:imp-detail}}

\subsection{Model and hyperparameters}
In this section, we provide a summary of the architecture and hyperparameters used in the Encoder-Decoder Control Transformer. Our model consists of a ViT encoder, a TokenLearner, and a Control Transformer. The Control Transformer we use is composed of 8 causal attention layers with 8 attention heads, 8 cross-attention layers with 8 attention heads, and an embedding dimension of 512. The ViT encoder is ViT-B/16 and we load pretrained weights from MultiMAE~\cite{bachmann2022multimae}. Instead of using mean pooling and a linear projection layer, we employ a TokenLearner that subsamples the 196 patch tokens output by the ViT encoder to 8 tokens, which are then passed to the Transformer decoder layers.

For both Phase \Romannum{1} and Phase \Romannum{2}, we utilize the default AdamW optimizer~\cite{loshchilovdecoupled}. For Phase \Romannum{1}, the learning rate and batch size are set to 5e-5 and 16, respectively, while for Phase \Romannum{2}, they are set to 1e-4 and 128, respectively. Additionally, a context length of 6 is used in all models for both training and execution. Phase \Romannum{1} has 175M trainable parameters while Phase \Romannum{2} has 51.1M trainable parameters. All models are trained for 10 epochs in Phase \Romannum{1}, and 10 epochs for Phase \Romannum{2}, with additional training details provided in Sec.~\ref{supp:train_detail}.







\subsection{Baselines architecture}

We summary the differences between \ours and Baselines in Table ~\ref{tab:study-baselines}. The details of Baselines are listed below.

\begin{table*}[t!]
    \centering
    \small 
    \begin{tabular}{l|lll}
    \toprule
                                         & Training objectives                  & Model structure                     & \duelph  \\ \hline
        \multirow{2}{*}{\ours}        & Phase \Romannum{1}: Self-superv.     &  Phase \Romannum{1}.:\ourmod                 & \cmark  \\
                                         &  Phase \Romannum{2}: IL-prompt      &  Phase \Romannum{2}: +\xatten                              &  \\ \hline
        \texttt{IL-only}           & IL-prompt      & \ourmod +\xatten           & \xmark  \\  \hline
        \texttt{SMART-only}  &  Self-superv. prompt            & \ourmod +\xatten                      & \xmark  \\  \hline
        \texttt{Jointly}                  & Self-superv. + IL-prompt                         & \ourmod +\xatten                         & \xmark \\ \hline
       \texttt{GATO-CT}                  & IL-prompt                   & \ourmod                      & \xmark \\ \hline
        \texttt{GATO*}              & IL-prompt                        & GATO~\cite{reed2022generalist}                         & \xmark \\ 
       
        \bottomrule
    \end{tabular}
    \caption{Comparisons of different baselines.}
    \label{tab:study-baselines}
\end{table*}

\subsubsection{\texttt{IL-only}, \texttt{SMART-only} and \texttt{Jointly}. }

The models \texttt{IL-only}, \texttt{SMART-only}, and \texttt{Jointly} all employ the same architecture as \ours during Phase \Romannum{2}. This architecture encompasses a Transformer encoder (with State and Action tokenizers), a decoder, and a \xatten layer. The only difference between them is the modification of the training objectives and phase.

The \texttt{IL-only} model focuses solely on prompt-conditioned imitation learning during its training process. In contrast, the \texttt{SMART-only} model leverages SMART training objectives in a purely self-supervised learning context with prompt-conditioning. The \texttt{Jointly} model synthesizes these methods, employing both SMART objectives and prompt-conditioned imitation learning loss in its comprehensive training strategy.

\subsubsection{\texttt{GATO-CT} and \texttt{GATO*}}

\setlength{\parindent}{2em}\textbf{\texttt{GATO*}: }GATO~\cite{reed2022generalist} (In this paper, we use \textbf{\texttt{GATO*}} to denote) is decode-only model, which imitates expert demonstrations from a vast dataset by prompting the model with the state and action subsequence. This model has 1.18 billion parameters and was trained on massive datasets, including 94.6k episodes from Metaworld. We include its reported performance on the Metaworld benchmark for reference. 

\begin{figure}
    \centering
    \includegraphics[width=.4\textwidth]{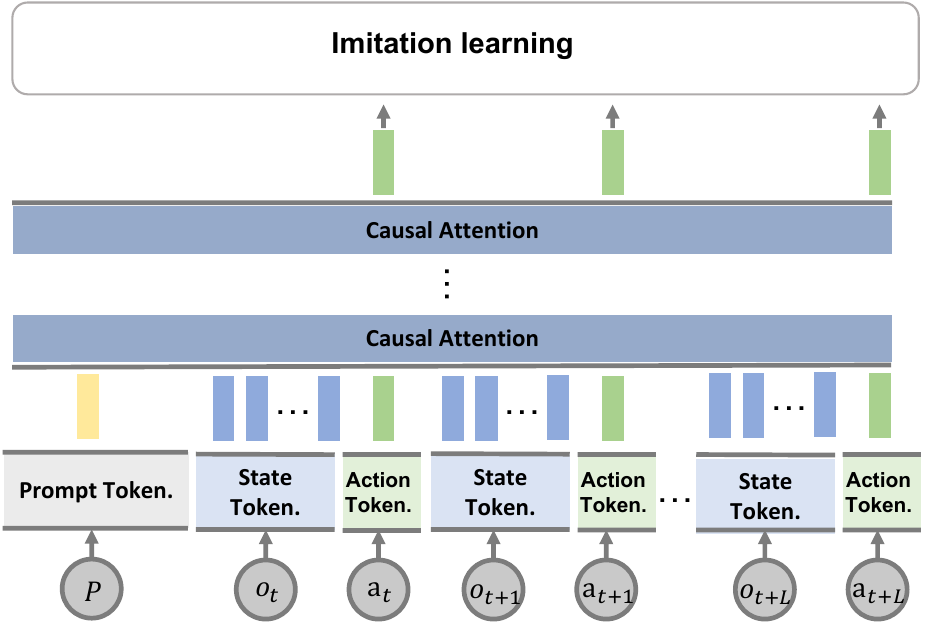}
    \caption{The architecture diagram of \texttt{GATO-CT}.}
    \label{fig:supp-gato}
\end{figure}

\textbf{\texttt{GATO-CT}: }For a fair comparison, we used the same base model architecture (\ourmod), but replaced our \xatten-based prompting approach with their proposed prefix prompting approach, denote as \textbf{\texttt{GATO-CT}}. Similar to \texttt{IL-only}, only imitation learning loss is used to predict future actions, But replace the \xatten module with one that prefixes the model with prompt token. Details are provided in Fig. ~\ref{fig:supp-gato}.

\subsection{Data collection}{\label{sup:data_collect}}

\textbf{Habitat.}
We collect shortest path episodes sampled from each of the 72 Gibson~\cite{xia2018gibson}, 61 mp3d~\cite{chang2017matterport3d} and 800 hm3d~\cite{ramakrishnan2021hm3d:habitat_benchmark} training scenes. These demonstrations are generated by greedily fitting actions to follow the geodesic shortest path to the nearest navigable goal object viewpoint. We hold out 5 Gibson scenes (hominy, Goffs, Hillsdale, Micanopy, and Rosser) for the experiments of out-of-distribution. The data we collected included RGB images ($3 \times 224 \times 224$), goal, and actions. We collected about 1000 episodes for each scene. Then, we divided the dataset and created the following dataset based on their intended purposes.
 \begin{itemize}
    \item \textbf{Habitat 50k.} We select all scenes of the Habitat dataset, and randomly sample about 50 episodes pre scene from the Habitat dataset. This data has 50K episodes and about $\sim$3.26M interaction steps. The dataset is used in Phase \Romannum{1} and Phase \Romannum{2} training.
    \item \textbf{Habitat 10k.} We randomly select 10 scenes of Habitat scenes, and randomly sample 1000 episodes pre scene from the Habitat dataset. This data has 10K episodes and about $\sim$0.54M interaction steps. The dataset is used to train the model in Phase \Romannum{2} of the ablation study.
 
    \item \textbf{Out-of-distribution tasks.} We select 5 Gibson scenes (”Goffs”, ”Hominy”, ”Hillsdale”, ”Micanopy”, and ”Rosser”) held out, and randomly sample 10, 100, and 1000 episodes pre scenesfrom the Habitat dataset.
\end{itemize}
\textbf{Metaworld.}
We collected data for all tasks in the MT50~\cite{yu2020meta} using scripted policies, which allowed us to generate expert demonstrations across an unlimited number of environment seeds. The data we collected included RGB images (3×224×224) rendered by the physical simulator, physics engine states, and actions. We collected 2000 episodes for each tasks. We use 45 tasks in the ML45 for Phase \Romannum{1}, and hold out other 5 tasks (hand-insert-v2, door-lock-v2, door-unlock-v2, box-close-v2 and bin-picking-v2) for the experiments of out-of-distribution. Then, we divided the dataset and created the following dataset based on their intended purposes.
\begin{itemize}
    \item \textbf{ML45.} We select 45 training tasks of ML45 in Metaworld, and randomly sample 1000 episodes pre task from the Metaworld dataset. This data has 45K episodes and about $\sim$3.40M interaction steps. The dataset is used in Phase \Romannum{1} and Phase \Romannum{2} training.
    \item \textbf{ML10.} We select 10 training tasks of ML10 in Metaworld, and randomly sample 1000 episodes pre task from the Metaworld dataset. This data has 10K episodes and about $\sim$0.79M interaction steps. The dataset is used to train the model in Phase \Romannum{2} of the ablation study.
    \item \textbf{Out-of-distribution tasks.} We select 5 test tasks of ML45 in Metaworld (”hand-insert-v2”, ”door-unlock-v2”, ”door-lock-v2”, ”box-close-v2”, and ”bin-picking-v2”), and randomly sample 10, 100, and 1000 episodes pre task from the Metaworld dataset.
\end{itemize}

\subsection{Training detail}{\label{supp:train_detail}}

\textbf{Phase \Romannum{1}.}
In Phase \Romannum{1}, the entire model, except for the cross-attention layers (\xatten), is trained using a self-supervised training objective on the ML45 dataset. 
 
\textbf{Phase \Romannum{2}.}
In Phase \Romannum{2}, we freeze the model encoder and only finetune a small part of the model, namely the Control Transformer, using imitation learning based on prompts. To encode the prompts, we use the CLIP encoder (CLIP/ViT-B/16)~\cite{radford2021learning:clip} and denote the resulting prompt sequence as $P$. The output sequence from each cross-attention layer is computed by softmax$(\frac{q_{H}k_p^T}{\sqrt{d}})v_P$,  where $H$ is the sequence of episodes and $d$ is the embedding dimension. In ablation study, we use Habitat 10K and ML10 datasets for Phase \Romannum{2} training dataset, while for the other experiments we use the Habitat 50K and ML45 datasets as training data for Phase \Romannum{2}.

\textbf{Out-of-distribution tasks.} 
In Sec.~\ref{ssec:ablate-ood}, we use \texttt{\ours}, \texttt{IL-only}, and \texttt{Scratch} for out-of-distribution tasks. \texttt{\ours} and \texttt{IL-only} model are trained beforehand and further finetuned with few-shot demonstrations. \texttt{Scratch} refers to the model that is trained on few-shot demonstrations from randomly initialized model weights.  We randomly select 10, 100 and 1000 episodes for few-shot learning. We use batch size bs = 64 and lr = 1e-4.  We train all models for 10000 gradient steps. The data for the out-of-distribution tasks are generated in the same way as we did in Sec.~\ref{sup:data_collect}.

\textbf{Ablation study.}
 In Sec.~\ref{ssec:ablate}, we use  Phase \Romannum{1} model pretrained on Habitat 50k and ML45 datasets. And the training parameters were the same as in Phase \Romannum{2} except for the change in ablation condition and datasets.


\subsection{Evaluation detail}{\label{supp:eval_detail}}

\textbf{Habitat. }Habitat is an immersive navigation task that provides a visually realistic environment. Our experiments are focused on the ImageNav task, in which the agent navigates towards a target position based on a goal image. The agent should stop within 1000 steps and reach a distance of 1m from the target image. To conduct our evaluation, we chose 4 scenes (Convoy, Beach, Cooperstown and Eagerville). We hold out 5 Gibson scenes (hominy, Goffs, Hillsdale, Micanopy, and Rosser) for the experiments of out-of-distribution tasks. For each scene, we randomly select three difficulty levels based on path length (EASY: 1.5-3m, MEDIUM: 3-5m, and HARD: 5-10m), resulting in a total of 300 episodes per scene. The metrics of the Habitat benchmark are listed below:
\begin{itemize}
    \item \textbf{Success Rate(SR) and Success weighted by Path Length(SPL).} The success rate(SR) and success eighted by Path Length(SPL), proposed by~\cite{anderson2018evaluation}, are estimated over 100 episodes on 4 scenes with 3 difficulty levels per scene, for a total of 1200 episodes per seed.
\end{itemize}

\textbf{Metaworld. }Metaworld is a benchmark of 50 diverse simulated manipulation tasks. We select 45 training tasks on ML45 for the experiments of evaluation, and hold out 5 test tasks (”hand-insert-v2”, ”door-unlock-v2”, ”door-lock-v2”, ”box-close-v2”, and ”bin-picking-v2”) for the experiments of out-of-distribution. The metrics of the Metaworld benchmark are listed below.
\begin{itemize}
    \item \textbf{Success Rate(SR)}. We refer to the evaluation method in Metaworld~\cite{yu2020meta}. The success rate is estimated over 10 seeds per task.
    \item \textbf{Expert Score}. The expert score is a measure of the difference between the performance of agents and experts, and is calculated as the ratio of the return obtained by agents to the expert return. We use the same expert return calculation method as GATO~\cite{reed2022generalist}.
    $$
    \max _{j \in[0,1, \ldots, N-W]}\left(\sum_{i=j}^{j+L-1} \frac{R_i}{W}\right)
    $$
where $N$ it the total number of collected episodes for the task, $W$ is the window size, and $R_i$ is the total return for episode $i$.

\end{itemize}

\section{More experiments} \label{app:more-exp}

\subsection{Comparisons of varying context length }

\begin{figure}[ht!]
    \centering
    \hspace{4em}
    \begin{subfigure}[t]{0.3\columnwidth}
        \centering
        \resizebox{1.1\columnwidth}{!}{
\begin{tikzpicture}


\definecolor{color1}{rgb}{0.466666666666667,0.533333333333333,0.6}
\definecolor{color0}{rgb}{0.67843137254902,0.847058823529412,0.901960784313726}
\definecolor{color2}{rgb}{1,0.870588235294118,0.67843137254902}
\definecolor{color3}{rgb}{1,0.752941176470588,0.796078431372549}

\begin{axis}[
axis line style={white!80!black},
legend cell align={left},
legend style={
  fill opacity=0.8,
  draw opacity=1,
  text opacity=1,
  at={(0.2,0.2)},
  anchor=center,
  draw=none,
  fill=none
},
tick align=outside,
tick pos=left,
x grid style={white!80!black},
xlabel={\Large{Expert Score/\%}},
xmajorgrids,
xmin=-5, xmax=105,
xtick style={color=white!15!black},
y grid style={white!80!black},
y label style={at={(axis description cs:0.08,0.5)},anchor=south},
ylabel={\large{Number of Tasks}},
ymajorgrids,
ymin=-0.15, ymax=47.15,
ytick style={color=white!15!black}
]
\addplot [ultra thick, color0, mark=*, mark size=3, mark options={solid}]
table {%
0 45
10 45
20 43
30 41
40 39
50 39
60 37
70 37
80 32
90 27
100 18
};
\addlegendentry{context length=6}

\addplot [ultra thick, color1, mark=*, mark size=3, mark options={solid}]
table {%
0 45
10 42
20 42
30 40
40 39
50 38
60 35
70 30
80 25
90 23
100 10
};
\addlegendentry{context length=8}

\addplot [ultra thick, color2, mark=*, mark size=3, mark options={solid}]
table {%
0 45
10 44
20 42
30 39
40 37
50 37
60 35
70 32
80 31
90 30
100 15
};
\addlegendentry{context length=12}

\addplot [ultra thick, color3, mark=*, mark size=3, mark options={solid}]
table {%
0 45
10 44
20 43
30 42
40 41
50 40
60 36
70 34
80 30
90 25
100 15
};
\addlegendentry{context length=16}
29 2 3 2 2 1 1 0 3 0 2

\end{axis}

\end{tikzpicture}} 
        \label{sfig:meta45_cl_es}
    \end{subfigure}
    \hfill
    \begin{subfigure}[t]{0.3\columnwidth}
        \centering
        \resizebox{1.1\columnwidth}{!}{
\begin{tikzpicture}


\definecolor{color1}{rgb}{0.466666666666667,0.533333333333333,0.6}
\definecolor{color0}{rgb}{0.67843137254902,0.847058823529412,0.901960784313726}
\definecolor{color2}{rgb}{1,0.870588235294118,0.67843137254902}
\definecolor{color3}{rgb}{1,0.752941176470588,0.796078431372549}

\begin{axis}[
axis line style={white!80!black},
legend cell align={left},
legend style={
  fill opacity=0.8,
  draw opacity=1,
  text opacity=1,
  at={(0.2,0.2)},
  anchor=center,
  draw=none, 
  fill=none
},
tick align=outside,
tick pos=left,
x grid style={white!80!black},
xlabel={\Large{Success Rate}},
xmajorgrids,
xmin=-5, xmax=105,
xtick style={color=white!15!black},
y grid style={white!80!black},
y label style={at={(axis description cs:0.08,0.5)},anchor=south},
ylabel={\large{Number of Tasks}},
ymajorgrids,
ymin=-2.25, ymax=47.25,
ytick style={color=white!15!black}
]
\addplot [ultra thick, color0, mark=*, mark size=3, mark options={solid}]
table {%
0 45
10 42
20 40
30 40
40 37
50 37
60 36
70 36
80 34
90 31
100 28
};
\addlegendentry{context length=6}

\addplot [ultra thick, color1, mark=*, mark size=3, mark options={solid}]
table {%
0 45
10 40
20 37
30 37
40 36
50 34
60 34
70 33
80 29
90 28
100 23
};
\addlegendentry{context length=8}
\addplot [ultra thick, color2, mark=*, mark size=3, mark options={solid}]
table {%
0 45
10 41
20 40
30 37
40 36
50 36
60 35
70 32
80 29
90 27
100 21
};
\addlegendentry{context length=12}

\addplot [ultra thick, color3, mark=*, mark size=3, mark options={solid}]
table {%
0 45
10 42
20 39
30 38
40 37
50 36
60 32
70 30
80 28
90 28
100 25
};
\addlegendentry{context length=16}
\end{axis}

\end{tikzpicture}}
        \label{sfig:meta45_cl_sr}
    \end{subfigure}
    \hspace{4em}
    \vspace{-2em}
    \caption{Comparisons of varying \textbf{context length} on \textit{MetaWorld 45 tasks} on Percentage of Expert Score (PES) (left) and Success Rate (SR) (right).}
    \label{fig:g-meta45_cl}
    \vspace{-1em}
\end{figure}
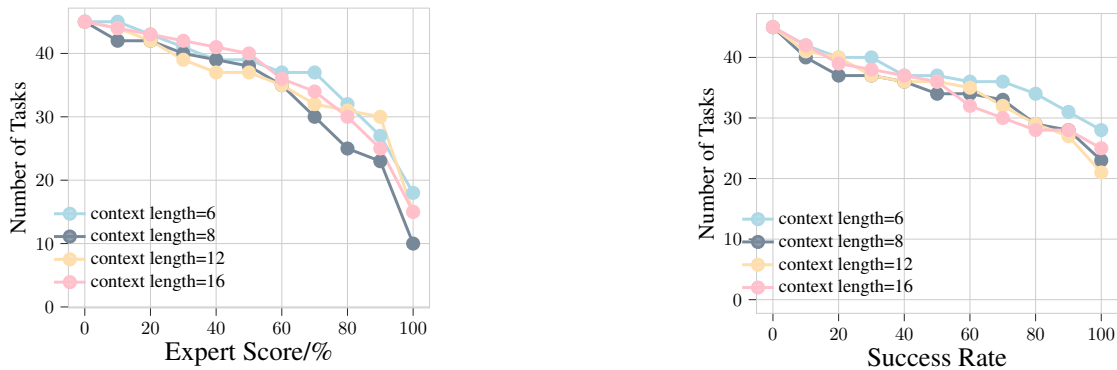

We conducted experiments on different context lengths, as illustrated in Fig.~\ref{fig:g-hab-cl} and Fig.~\ref{fig:g-meta45_cl}. 
On the navigation tasks in Habitat, long-range temporal dependencies are important for decision-making. As a result, the model's performance is improved progressively as the length of the context increases, as shown in Fig.~\ref{fig:g-hab-cl}. 
On the other hand, we observed that setting the context length to 6 leads to better performance on the Metaworld dataset, as demonstrated in Fig.~\ref{fig:g-meta45_cl}. Therefore, we choose a context length of 6 as means of balancing performance and compute cost. However, if one seeks to capture long-term temporal dependence, increasing the context length may be necessary.

\begin{figure}[ht!]
    \centering
    \hspace{4em}
    \begin{subfigure}[t]{0.3\columnwidth}
        \centering
        \resizebox{1.1\columnwidth}{!}{
\begin{tikzpicture}

\definecolor{color1}{rgb}{0.466666666666667,0.533333333333333,0.6}
\definecolor{color0}{rgb}{0.67843137254902,0.847058823529412,0.901960784313726}
\definecolor{color2}{rgb}{1,0.870588235294118,0.67843137254902}
\definecolor{color3}{rgb}{1,0.752941176470588,0.796078431372549}

\begin{axis}[
width=10cm,
height=7cm,
axis line style={white!80!black},
tick align=outside,
tick pos=left,
x grid style={white!80!black},
xmin=0.065, xmax=0.9,
xtick style={color=white!15!black},
xtick={0.2,0.4,0.6,0.8},
xticklabels={\textbf{6},\textbf{8},\textbf{12},\textbf{16}},
xlabel={\large{Context Length}},
y grid style={white!80!black},
ylabel={\large{Success Rate}},
ymajorgrids,
ymin=0, ymax=0.16,
scaled y ticks=false,
ytick style={color=white!15!black},
ytick={0.0,0.04,0.08,0.12,0.16},
yticklabels={0.0,0.04,0.08,0.12,0.16},
]
\draw[draw=none,fill=color0] (axis cs:0.15,0) rectangle (axis cs:0.25,0.1239);
\draw[draw=none,fill=color1] (axis cs:0.35,0) rectangle (axis cs:0.45,0.1175);
\draw[draw=none,fill=color2] (axis cs:0.55,0) rectangle (axis cs:0.65,0.131944444);
\draw[draw=none,fill=color3] (axis cs:0.75,0) rectangle (axis cs:0.85,0.140833333);
\end{axis}

\end{tikzpicture}} 
        \label{sfig:hat_cl_sr}
    \end{subfigure}
    \hfill
    \begin{subfigure}[t]{0.3\columnwidth}
        \centering
        \resizebox{1.1\columnwidth}{!}{
\begin{tikzpicture}

\definecolor{color1}{rgb}{0.466666666666667,0.533333333333333,0.6}
\definecolor{color0}{rgb}{0.67843137254902,0.847058823529412,0.901960784313726}
\definecolor{color2}{rgb}{1,0.870588235294118,0.67843137254902}
\definecolor{color3}{rgb}{1,0.752941176470588,0.796078431372549}

\begin{axis}[
width=10cm,
height=7cm,
axis line style={white!80!black},
tick align=outside,
tick pos=left,
x grid style={white!80!black},
xmin=0.065, xmax=0.9,
xtick style={color=white!15!black},
xtick={0.2,0.4,0.6,0.8},
xticklabels={\textbf{6},\textbf{8},\textbf{12},\textbf{16}},
xlabel={\large{Context Length}},
y grid style={white!80!black},
ylabel={\large{Success Rate}},
ymajorgrids,
ymin=0, ymax=0.16,
scaled y ticks=false,
ytick style={color=white!15!black},
ytick={0.0,0.04,0.08,0.12,0.16},
yticklabels={0.0,0.04,0.08,0.12,0.16},
]
\draw[draw=none,fill=color0] (axis cs:0.15,0) rectangle (axis cs:0.25,0.1091);
\addlegendimage{ybar,ybar legend,draw=none,fill=color0}
0.		

\draw[draw=none,fill=color1] (axis cs:0.35,0) rectangle (axis cs:0.45,0.106083333);
\addlegendimage{ybar,ybar legend,draw=none,fill=color1}

\draw[draw=none,fill=color2] (axis cs:0.55,0) rectangle (axis cs:0.65,0.102855556);
\addlegendimage{ybar,ybar legend,draw=none,fill=color2}

\draw[draw=none,fill=color3] (axis cs:0.75,0) rectangle (axis cs:0.85,0.112);


\end{axis}

\end{tikzpicture}}
        \label{sfig:hat_cl_spl}
    \end{subfigure}
    \hspace{4em}
    \vspace{-1.5em}
    \caption{Comparison of varying \textbf{context length} on \textit{Habitat},and compare agents by Success Rate (SR) (left) and Success weighted by Path Length (SPL) (right).}
    \label{fig:g-hab-cl}
    \vspace{-0.5em}
\end{figure}
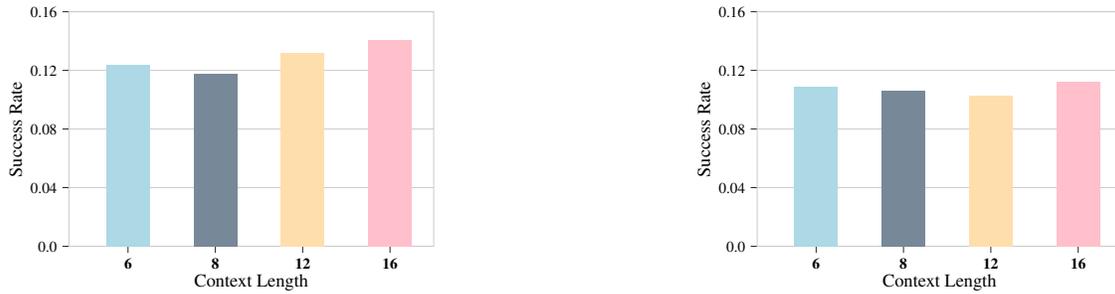

\subsection{Comparison with vision tokenization}

We conducted a set of experiments to demonstrate the effectiveness of the multi-state tokens module (i.e., TokenLearner), by comparing it with a single-state tokens module that uses mean pooling and a linear projection layer to convert patch tokens to one token. In contrast to other ablation studies, we trained both Phase \Romannum{1} and Phase \Romannum{2} on the ML45 and Habitat 50k datasets. The results in Table~\ref{tab:ablate-token} indicate a minor difference between the two on habitat, but an average success rate difference of approximately 0.09 on ML45. These findings support our expectation that multi-state tokens can extract additional information from the encoder to improve decision-making and enhance overall learning performance.

\begin{table}[h!]
    \centering
    \small 
    \begin{tabular}{c|lll}
    \toprule
    tokenization  & Habitat      & Metaworld   \\    \hline
     multi-state token.  & 0.1239 & 0.802              \\
    single-state token.   & 0.1217   & 0.713\\
    \bottomrule
    \end{tabular}
    \caption{Comparisons of tokenization methods on \textit{Habitat} and \textit{MetaWorld 45} tasks, measured by Success Rate (SR).}
    \label{tab:ablate-token}
\end{table}

\subsection{Prompt conditioning discussion}
\textbf{Implementation details. }In Sec.~\ref{ssec:ablate-prompt}, we conducted an ablation study by comparing two prompt conditioning approaches: prefix and \xatten prompting. The prefix approach is a conventional prompting method that splices the prompt sequences in front of the token sequences, which are directly fed into the Transformer decoder layers. In contrast, \xatten prompting uses a cross-attention layer to fuse the prompt sequences and token sequences together. We utilized the base model that was pretrained on Habitat 50 and ML45 after Phase \Romannum{1}. We use Habitat 10K and ML10 datasets for Phase II training dataset. 

\textbf{Discussion. }In the experiments discussed in Sec.\ref{ssec:ablate-prompt}, it was found that \xatten prompting outperforms prefix prompting. This suggests that the cross attention mechanism is effective in establishing a strong connection between prompt and token sequences, which has also been demonstrated in other recent works, such as Vima\cite{jiang2022vima} and Stable Diffusion~\cite{rombach2022high}. One potential limitation of prefix prompting is that the prompt token sequence may be too short to attract sufficient attention from the attention mechanism, leading to suboptimal performance. To address this, future research may explore alternative encoding methods for prompts that can better capture the information necessary for guiding the model's output.



\subsection{Attention visualization}
\begin{figure*}[h]
    \centering
    \includegraphics[width=.9\textwidth]{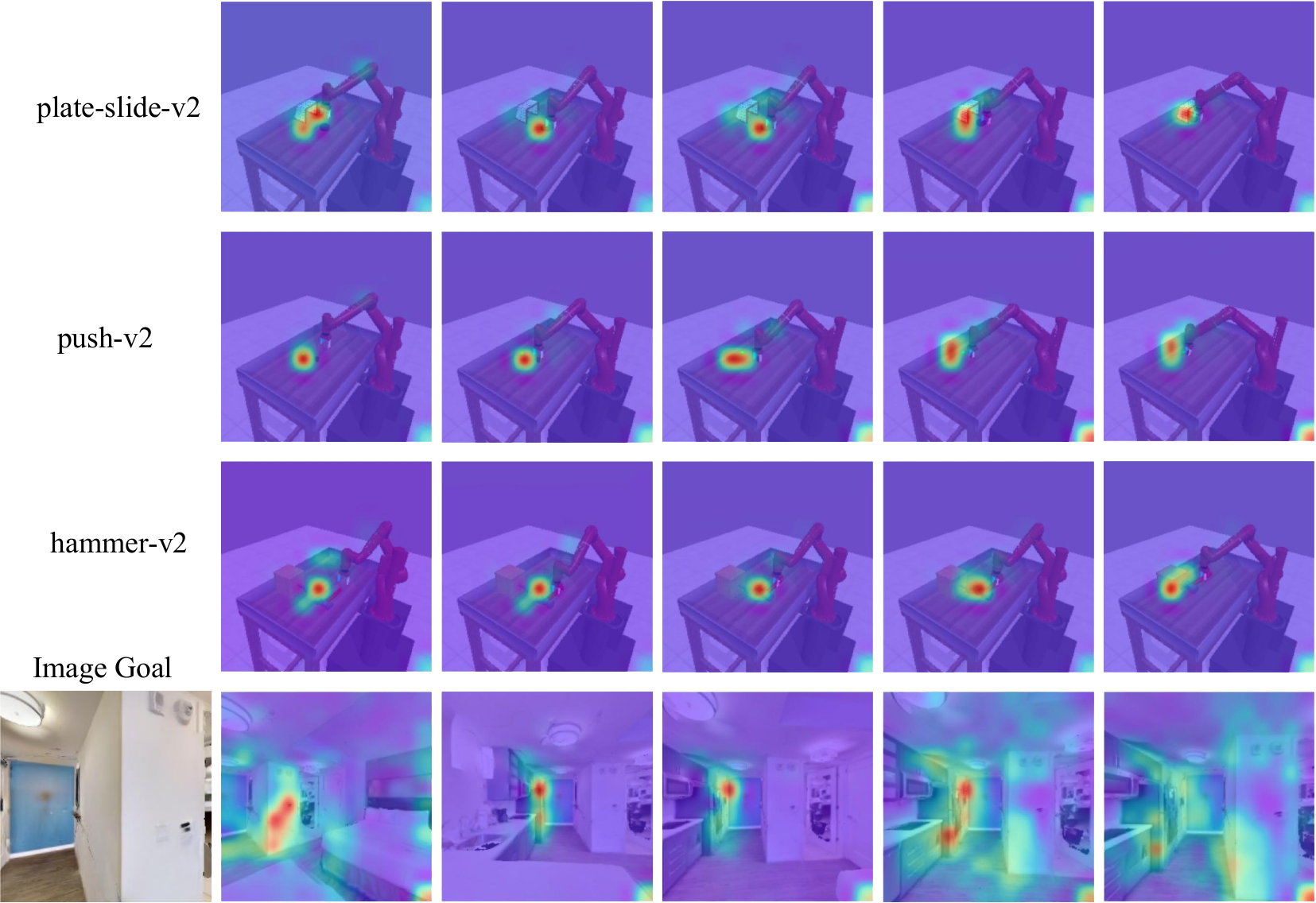}
    \caption{More attention map visualization. On Metaworld, the attention maps show that the model focuses on the
location of the item being manipulated, and then interacts with the corresponding item to complete the task. On Habitat, \ours focuses more on exploring the scene and then locating the goal.}
    \label{fig:supp_attn_map}
\end{figure*}

In Figure~\ref{fig:supp_attn_map}, we provide additional attention maps that reveal how \ours tends to focus on the object being manipulated, as well as its surrounding context and relevant visual cues (such as ``plate-slide-v2", ``push-v2", and ``hammer-v2") when performing manipulation tasks in MetaWorld. Furthermore, the attention maps show that the model focuses on the location of the item being manipulated, and then interacts with the corresponding item to complete the task. In Habitat, our model (\ours) focuses more on exploring the scene and then locating the goal, as illustrated in Figure~\ref{fig:supp_attn_map}. The attention maps demonstrate that \ours quickly identified the location of the goal image at the outset. Despite that there are obstacles blocking the shortest path, \ours was able to eventually reach the goal.

\subsection{Performance on each tasks}
We show  the detailed results of all models on Metaworld and Habitat in Table~\ref{tab:task-detail-hab}, Table~\ref{tab:task-detail-meta-1}, and Table~\ref{tab:task-detail-meta-2}. Specifically, Table~\ref{tab:task-detail-hab} presents the Habitat results, while Table~\ref{tab:task-detail-meta-1} and Table~\ref{tab:task-detail-meta-2} present the Metaworld results.

\clearpage
 \begin{table*}[h]
    \centering
    \small 
    \begin{tabular}{c|llllllll}
    \toprule
    &\multicolumn{2}{c}{\ours} &  \multicolumn{2}{c}{\ours/single} \\ 
    scene  & SR      & SPL  & SR    & SPL    \\    \hline
    Convoy(easy)  &  0.143$\pm$ 0.021 & 0.124$\pm$0.017  & 0.160$\pm$0.026  & 0.115$\pm$0.027 \\ 
    Convoy(medium)  & 0.150$\pm$0.017   & 0.144$\pm$0.020  &0.157$\pm$0.006&	0.126$\pm$0.009  \\
    Convoy(hard)  &  0.067$\pm$0.025   &0.062$\pm$0.029 &0.173$\pm$0.042&	0.162$\pm$0.039  \\
    Beach(easy)  &0.170$\pm$0.044 & 0.144$\pm$0.045 &0.170$\pm$0.053 &0.133$\pm$0.036  \\
    Beach(medium)  & 0.157$\pm$0.015  & 0.146$\pm$0.012 &0.200$\pm$0.050&	0.165$\pm$0.050 \\
    Beach(hard)  & 0.133$\pm$0.015  & 0.128$\pm$0.014&0.227$\pm$0.045&	0.215$\pm$0.047  \\
    Cooperstown(easy)  &0.147$\pm$0.021 & 0.122$\pm$0.030 &0.180$\pm$0.070&0.141$\pm$0.058 \\
    Cooperstown(medium)  &  0.110$\pm$0.035   & 0.103$\pm$0.027 &0.190$\pm$0.046&	0.175$\pm$0.044   \\
    Cooperstown(hard)  &0.073$\pm$0.021  & 0.070$\pm$0.020&0.140$\pm$0.046&	0.132$\pm$0.043  \\
    Eagerville(easy)  & 0.113$\pm$0.025 & 0.079$\pm$0.027 &0.087$\pm$0.015&0.054$\pm$0.003 \\
    Eagerville(medium)  &  0.127$\pm$0.029  & 0.106$\pm$0.020 &0.107$\pm$0.025&	0.087$\pm$0.019   \\
    Eagerville(hard)  & 0.097$\pm$0.046  & 0.081$\pm$0.043&0.147$\pm$0.021&	0.130$\pm$0.018\\ \hline
     & \multicolumn{2}{c}{Jointly}& \multicolumn{2}{c}{Jointly/single}\\ 
    scene  & SR     & SPL  & SR    & SPL   \\    \hline
    Convoy(easy)  &  0.130$\pm$0.020    &0.122$\pm$0.023  & 0.173$\pm$0.031  & 0.120$\pm$0.020 \\ 
    Convoy(medium)  & 0.080$\pm$0.000   & 0.076$\pm$0.002  &0.143$\pm$0.031 &	0.111$\pm$0.035  \\
    Convoy(hard)  & 0.050$\pm$0.010   &0.049$\pm$0.011 &0.177$\pm$0.015 &	0.160$\pm$0.017 \\
    Beach(easy)   &  0.137$\pm$0.021 &0.115$\pm$0.013 &0.220$\pm$0.056 & 0.156$\pm$0.026 \\
    Beach(medium)  & 0.093$\pm$0.021   & 0.086$\pm$0.019 &0.230$\pm$0.030 & 0.172$\pm$0.015 \\
    Beach(hard)  & 0.057$\pm$0.021  & 0.054$\pm$0.021 &0.147$\pm$0.042 &	0.115$\pm$0.018 \\
    Cooperstown(easy)  & 0.117$\pm$0.035 &0.106$\pm$0.034 &0.173$\pm$0.021&0.139$\pm$0.027\\
    Cooperstown(medium)  & 0.077$\pm$0.006   & 0.069$\pm$0.007 &0.240$\pm$0.026 &	0.219$\pm$0.027 \\
    Cooperstown(hard)   & 0.080$\pm$0.020 &0.076$\pm$0.022 &0.183$\pm$0.032 &	0.167$\pm$0.032  \\
    Eagerville(easy)  & 0.137$\pm$0.035 & 0.116$\pm$0.035 &0.100$\pm$0.010 & 0.058$\pm$0.002  \\
    Eagerville(medium)   & 0.097$\pm$0.015  &0.085$\pm$0.016 &0.180$\pm$0.030 & 0.112$\pm$0.021  \\
    Eagerville(hard) & 0.067$\pm$0.023  & 0.061$\pm$0.022 &0.193$\pm$0.040 &	0.150$\pm$0.020  \\ \hline
    &\multicolumn{2}{c}{IL-only } &  \multicolumn{2}{c}{IL-only/single}  \\ 
    scene   & SR      & SPL  & SR    & SPL      \\    \hline
    Convoy(easy)  & 0.113$\pm$0.006    &0.110$\pm$0.006  & 0.110$\pm$0.010  & 0.09$\pm$0.011  \\ 
    Convoy(medium)  &0.040$\pm$0.026  & 0.039$\pm$0.026  &0.037$\pm$0.025 &	0.030$\pm$0.021 \\
    Convoy(hard)  & 0.027$\pm$0.006   &0.027$\pm$0.006 &0.053$\pm$0.012 &	0.043$\pm$0.011 \\
    Beach(easy)  & 0.103$\pm$0.055 & 0.095$\pm$0.051 &0.070$\pm$0.010 & 0.052$\pm$0.010  \\
    Beach(medium)  &0.053$\pm$0.035   & 0.050$\pm$0.033 &0.050$\pm$0.010 & 0.035$\pm$0.0135\\
    Beach(hard)  & 0.030$\pm$0.017  & 0.027$\pm$0.015 &0.033$\pm$0.012 &	0.027$\pm$0.010\\
    Cooperstown(easy)  & 0.080$\pm$0.010 & 0.074$\pm$0.010 & 0.103$\pm$0.015 &0.076$\pm$0.01   \\
    Cooperstown(medium)  &  0.043$\pm$0.012   & 0.041$\pm$0.023 &0.053$\pm$0.012 &	0.038$\pm$0.013  \\
    Cooperstown(hard)  &0.047$\pm$0.020 &0.045$\pm$0.009 & 0.050$\pm$0.000 &	0.039$\pm$0.006 \\
    Eagerville(easy)  & 0.067$\pm$0.025 & 0.064$\pm$0.025 &0.063$\pm$0.021 & 0.042$\pm$0.023  \\
    Eagerville(medium)  &  0.070$\pm$0.030  &0.054$\pm$0.018 &0.033$\pm$0.021 & 0.022 $\pm$0.01 \\
    Eagerville(hard)  & 0.047$\pm$0.021  & 0.040$\pm$0.024 & 0.033$\pm$0.021 &	0.026$\pm$0.014 \\ \hline
      & \multicolumn{2}{c}{SMART-only}& \multicolumn{2}{c}{SMART-only/single} \\ 
    scene   & SR     & SPL  & SR    & SPL   \\    \hline
    Convoy(easy)  & 0.113$\pm$0.006  & 0.088$\pm$0.003 & 0.007$\pm$0.006  & 0.007$\pm$0.006  \\ 
    Convoy(medium)   & 0.003$\pm$0.006  & 0.003$\pm$0.005   &0.0$\pm$0.0  & 0.0$\pm$0.0 \\
    Convoy(hard)  & 0.007$\pm$0.006  & 0.005$\pm$0.004   &0.0$\pm$0.0  & 0.0$\pm$0.0\\
    Beach(easy)   & 0.063$\pm$0.025  & 0.044$\pm$0.014 &0.007$\pm$0.012  &0.007$\pm$0.012  \\
    Beach(medium)  & 0.0$\pm$0.0  & 0.0$\pm$0.0 &0.0$\pm$0.0  & 0.0$\pm$0.0 \\
    Beach(hard)  & 0.010$\pm$0.010  &0.008$\pm$0.009  &0.0$\pm$0.0  & 0.0$\pm$0.0 \\
    Cooperstown(easy)   & 0.090$\pm$0.020 & 0.079$\pm$0.018 &0.013$\pm$0.006 & 0.013$\pm$0.006  \\
    Cooperstown(medium) & 0.010$\pm$0.010  & 0.007$\pm$0.006 &0.0$\pm$0.0  & 0.0$\pm$0.0 \\
    Cooperstown(hard)  & 0.003$\pm$0.006  &0.003$\pm$0.006  &0.0$\pm$0.0  & 0.0$\pm$0.0 \\
    Eagerville(easy)  & 0.087$\pm$0.006  & 0.076$\pm$0.003 & 0.017$\pm$0.006  & 0.016$\pm$0.005 \\
    Eagerville(medium)  & 0.023$\pm$0.006  &0.017$\pm$0.004  &0.0$\pm$0.0  & 0.0$\pm$0.0\\
    Eagerville(hard)  & 0.010$\pm$0.010  & 0.008$\pm$0.008  &0.0$\pm$0.0  & 0.0$\pm$0.0\\
    \bottomrule
    \end{tabular}

    \caption{performance on each tasks on Habitat}
    \label{tab:task-detail-hab}

\clearpage
\end{table*}
\begin{table*}[h]
    \centering
    \small 
    \begin{tabular}{c|llllllllllllllll}
    \toprule
    &\multicolumn{2}{c}{\ours} &  \multicolumn{2}{c}{\ours/single}&  \multicolumn{2}{c}{Jointly} &  \multicolumn{2}{c}{Jointly/single}  \\ 
    task  & SR     & return  & SR   & return & SR    & return  & SR    & return     \\    \hline
    assembly-v2    &  0.9    &1039.1  & 1.0 & 1276.0 & 0.0 & 198.8  & 0.0 & 477.0\\ 
    basketball-v2  & 0.3  & 400.0 & 0.5 & 621.7 & 0.0 & 10.6  & 0.0 & 42.4 \\
    button-press-topdown-v2  & 1.0   & 364.9 & 1.0 & 1175.4 & 0.6 & 113.9 & 0.6 &  45.2\\
    button-press-topdown-wall-v2)  & 1.0 & 0.0 & 1.0 & 0.0 & 0.7 &  55.2  & 1.0 & 39.5\\
    button-press-v2  & 1.0 & 347.3 & 1.0 & 357.6 & 0.7 & 318.2  &  0.5 & 326.2\\
    button-press-wall-v2 & 0.3 & 1373.7 & 1.0 & 1195.2 & 0.0 &  128.8  & 0.0 & 0.5\\
    coffee-button-v2  & 1.0 & 301.0 & 1.0 & 299.2 & 1.0 &  304.3  & 0.6 & 268.6\\
    coffee-pull-v2  &1.0  & 429.5 & 1.0 & 407.3 & 0.8 & 303.8  & 0.9 & 324.2\\
    coffee-push-v2  &1.0  & 443.6 & 1.0 & 509.0 & 0.0 & 30.3  & 0.2 & 66.9\\
    dial-turn-v2  & 1.0 & 1220.0 & 1.0 & 1203.3 & 0.0 &  61.6  & 0.3 & 17.7\\
    disassemble-v2  & 1.0  &615.3 & 1.0 & 553.4 & 0.0 & 220.9 & 0.0 & 213.2\\
    door-close-v2  & 1.0 & 946.0 & 1.0 & 754.9 & 0.1 & 2956.1 & 1.0 & 1286.4\\
    door-open-v2  & 1.0 & 1756.7 & 1.0 & 1775.4 & 0.1 & 696.5  & 0.9 & 1457.9\\
    drawer-close-v2  & 1.0 & 61.4 & 1.0 & 81.2 & 0.1 & 4.9  & 0.4 & 24.6\\
   drawer-open-v2  & 1.0 & 1965.0 & 1.0 & 1989.5 & 0.7 & 1517.8 & 0.5 & 1103.6\\
   faucet-open-v2  & 0.3 & 1693.4 & 1.0 & 2192.2 & 0.0 & 1388.9 & 0.0 &  1276.4\\
   faucet-close-v2  & 0.1 & 1624.0 & 0.0 & 1695.1 & 0.0 & 1856.1 & 0.2 & 2075.1\\
   hammer-v2  & 1.0 & 951.9 & 1.0 & 929.5 & 0.0 & 675.8 & 0.1 &  869.2\\
   handle-press-side-v2  & 1.0 & 839.9 & 1.0 & 808.8 & 0.7 & 307.2 & 1.0 & 877.9\\
   handle-press-v2  & 1.0 & 671.1 & 1.0 & 782.7 & 0.4 & 195.6  & 0.7 & 427.7\\
   handle-pull-side-v2  & 0.8 & 580.4 & 0.0 & 29.1 & 0.0 & 12.4 & 0.0 & 11.4\\
   handle-pull-v2  & 0.7 & 182.9 & 0.2 & 177.5 & 0.1 & 70.4  & 0.5 & 139.9\\
   lever-pull-v2  & 0.0 & 291.1 & 0.8 & 946.7 & 0.1 & 420.3  & 0.0 & 283.8\\
   peg-insert-side-v2  & 0.9 &  990.8 & 0.7 & 909.5 & 0.5 & 1413.7 & 0.3 & 1096.3\\
   pick-place-wall-v2  & 1.0 & 656.9 & 1.0 & 1698.2 & 0.0 & 0.1  & 0.0 & 14.0\\
   pick-out-of-hole-v2  & 0.0 & 261.1 & 0.0 & 362.4 & 0.0 & 25.9  & 0.0 & 12.3\\
   reach-v2  & 0.1 &  2507.2 & 0.0 & 2374.8 & 0.0 & 241.8  & 0.0 & 598.6\\
   push-back-v2  & 1.0 & 193.7 & 1.0 & 283.7 & 0.0 & 6.5 & 0.0 & 6.4\\
   push-v2  & 1.0 & 1264.1 & 1.0 & 1446.0 & 0.0 &  22.7 & 0.0 & 23.4\\
   pick-place-v2  & 0.7 & 608.4 & 1.0 & 303.0 & 0.0 & 7.2 & 0.0 & 19.2\\
   plate-slide-v2  & 1.0 & 1255.4 & 1.0 & 1214.2 & 0.0 & 269.7 & 0.2 & 417.6\\
   plate-slide-side-v2  & 1.0 & 1281.6 & 1.0 & 1278.6 & 0.0 & 143.1  & 0.2 & 397.8\\
   plate-slide-back-v2  & 1.0 & 1207.5 & 1.0 &  1170.6 & 0.8 & 909.0 & 0.4 & 618.6\\
   plate-slide-back-side-v2  & 1.0 & 1340.7 & 1.0 & 1321.9 & 0.1 & 499.6 & 0.6 & 782.1 \\
   peg-unplug-side-v2  & 0.8 & 343.9 & 1.0 & 344.0 & 0.1 & 109.2 & 0.3 & 122.0\\
   soccer-v2  & 0.0 &  336.4 & 0.0 & 329.6 & 0.1 & 180.3 & 0.0 & 164.3\\
   stick-push-v2  & 1.0 & 1328.5 & 1.0 & 1316.9 & 0.0 & 24.2  & 0.5 & 475.5\\
   stick-pull-v2  & 0.9 & 190.4 & 1.0 & 648.5 & 0.0 & 6.3 & 0.1 & 144.8 \\
   push-wall-v2  & 1.0 & 1387.4 & 1.0 & 1782.3 & 0.0 &  51.3  & 0.0 & 49.4\\
   reach-wall-v2  & 0.5 & 3151.8 & 0.0 & 4190.0 & 0.0 & 341.4 & 0.0 & 714.7 \\
   shelf-place-v2  & 0.8 & 752.2 & 0.9 & 864.4 & 0.0 & 0.0 & 0.0 & 0.1 \\
   sweep-into-v2  & 1.0 & 880.6 & 0.8 & 783.1 & 0.0 & 47.5 & 0.0 & 52.0\\
   sweep-v2  & 1.0 & 1346.6 & 1.0 & 1108.6 & 0.0 & 93.3 & 0.0 & 91.8\\
   window-open-v2  & 1.0 & 438.6 & 1.0 & 494.6 & 0.5 &  379.8 & 0.2 & 436.8\\
   window-close-v2  & 1.0 &  784.2 & 1.0 & 806.3 & 0.5 & 799.5 & 0.4 &  541.1\\ 
  
    \bottomrule
    \end{tabular}
    \caption{The detailed Metaworld ML45 results of the \ours, \ours/single, Jointly and Jointly/single on each tasks. }
    \label{tab:task-detail-meta-1}
\end{table*}

\clearpage
\begin{table*}[h]
    \centering
    \small 
    \begin{tabular}{c|llllllllllllllll}
    \toprule
    &  \multicolumn{2}{c}{IL-only} &  \multicolumn{2}{c}{IL-only/single} &
    \multicolumn{2}{c}{SMART-only} &  \multicolumn{2}{c}{SMART-only/single}\\
    & SR   & return & SR    & return & SR    & return & SR    & return \\  \hline
    assembly-v2    &  0.0   &252.1 & 0.0 & 169.1  & 0.0 & 197.2 & 0.0 & 189.0\\ 
    basketball-v2  & 0.0 & 13.9 & 0.0 & 5.1 & 0.0 & 2.0 & 0.0 & 1.3\\
    button-press-topdown-v2  & 0.0   & 194.3 & 0.0 &33.3 & 0.0 & 116.4& 0.0 & 0.1 \\
    button-press-topdown-wall-v2  & 0.6 & 88.6& 0.0 &  1.3 & 0.0 & 35.7& 0.0 & 14.5 \\
    button-press-v2  & 0.3  & 366.0 & 0.0 & 43.4 & 0.0 & 53.3& 0.0 & 45.3\\
    button-press-wall-v2 & 0.0 & 75.9 & 0.0 & 19.2 & 0.0 & 59.1& 0.0 & 25.2\\
    coffee-button-v2  & 1.0 & 301.0 & 0.0 & 38.6 & 0.6 & 297.3& 0.0 & 65.1\\
    coffee-pull-v2  & 0.9  & 365.8& 0.0 & 13.5 & 0.0 & 11.6 & 0.0 & 11.9\\
    coffee-push-v2  &0.0  &83.6& 0.0 & 13.1 & 0.0 & 10.9 & 0.0 & 5.1\\
    dial-turn-v2  &0.0  &19.3 & 0.0 & 6.7 & 0.0 &  4.4 & 0.0 &8.3\\
    disassemble-v2  &0.0  &210.9 & 0.0 & 206.6 & 0.0 & 204.8& 0.0 & 206.1 \\
    door-close-v2  &0.2  &2742.7 & 0.0 & 30.4 & 0.0 & 659.8& 0.2 & 327.4 \\
    door-open-v2   &0.4  &1169.6 & 0.0 & 169.6 & 0.0 & 212.3& 0.0 & 383.7 \\
    drawer-close-v2   &0.0  &0.0 & 1.0 & 71.3 & 0.0 & 2.3 & 0.0 & 0.0\\
   drawer-open-v2   &1.0  & 1976.5& 0.0 & 389.6 & 0.0 & 493.4& 0.0 & 389.9 \\
   faucet-open-v2   &0.1  &1547.4& 0.0 & 427.1& 0.0 & 490.4 & 0.0 & 302.0\\
   faucet-close-v2  &0.0  &1062.5& 0.0 & 453.2& 0.0 & 863.6 & 0.0 & 552.3\\
   hammer-v2   &0.0  &588.5& 0.0 & 528.0& 0.0 & 263.3 & 0.0 &  585.6\\
   handle-press-side-v2  &0.6  &235.4& 0.8 & 493.9& 0.0 & 26.3& 0.0 & 28.7 \\
   handle-press-v2  &0.0  &86.1 & 0.0 & 18.5& 0.0 & 34.3& 0.0 & 22.9 \\
   handle-pull-side-v2  &0.3  &12.1& 0.0 & 10.7& 0.0 & 2.1 & 0.0 & 2.5\\
   handle-pull-v2  &0.2  &82.5 & 0.0 & 6.4 & 0.0 & 14.4& 0.0 & 4.4\\
   lever-pull-v2   &0.0  &350.2& 0.0 & 24.2 & 0.0 & 125.0& 0.0 & 90.0\\
   peg-insert-side-v2  &0.9  &1289.4 & 0.0 & 2.2 & 0.0 & 2.4& 0.0 & 1.7\\
   pick-place-wall-v2  &0.0  &27.4& 0.0 & 0.0& 0.0 & 0.0 & 0.0 & 0.0\\
   pick-out-of-hole-v2   &0.0  &19.1 & 0.0 & 3.4 & 0.0 & 3.1& 0.0 & 1.2 \\
   reach-v2   &0.0  &195.0 & 0.0 & 122.3 & 0.0 & 144.5& 0.0 & 195.3\\
    push-back-v2 &0.0  &5.0 & 0.0 & 1.7 & 0.0 & 2.3& 0.0 &  1.7\\
   push-v2   &0.0  &21.8 & 0.0 & 10.2 & 0.0 & 3.8& 0.0 & 4.6\\
   pick-place-v2   &0.0  &6.7 & 0.0 & 3.0 & 0.0 & 2.3& 0.0 & 3.2\\
   plate-slide-v2   &0.3  &393.9 & 0.0 & 72.1& 0.0 & 97.2 & 0.0 & 44.2\\
   plate-slide-side-v2   &0.0  &45.8 & 0.2 & 419.0& 0.0 & 20.6 & 0.0 & 5.0\\
   plate-slide-back-v2   &0.7  &1027.7 & 0.0 & 43.3& 0.0 & 48.3& 0.0 &  21.8\\
   plate-slide-back-side-v2   &0.0  &200.1 & 0.0 & 1185.7& 0.0 & 25.5 & 0.0 & 25.6\\
   peg-unplug-side-v2   &0.2  &131.4 & 0.0 & 3.6 & 0.0 & 2.7& 0.0 &  2.8\\
   soccer-v2  &0.0  &21.0 & 0.0 & 38.1 & 0.0 & 3.4 & 0.0 &6.9\\
   stick-push-v2   &0.0  &16.7& 0.0 & 5.7 & 0.0 & 1.9& 0.0 & 3.1\\
   stick-pull-v2   &0.0  &6.6 & 0.0 & 5.8 & 0.0 &  2.2& 0.0 & 6.9\\
   push-wall-v2   &0.0  &24.4 & 0.0 & 18.0& 0.0 & 3.9& 0.0 & 5.0\\
   reach-wall-v2  &0.0  &558.7 & 0.0 & 305.2 & 0.0 & 159.9& 0.0 & 435.3\\
   shelf-place-v2   &0.0  &214.2& 0.0 &  0.0 & 0.0 & 0.0& 0.0 & 0.0\\
   sweep-into-v2   &0.0  &55.3 & 0.0 & 8.3& 0.0 & 10.7& 0.0 & 9.1\\
   sweep-v2   &0.0  &83.6 & 0.0 & 15.9 & 0.0 & 17.1 & 0.0 & 13.7\\
   window-open-v2   &1.0  &449.5& 0.0 & 101.4& 0.0 & 91.4& 0.0 & 92.7\\
   window-close-v2   &0.1  &462.9& 0.0 & 10.9& 0.0 & 374.4& 0.0 & 216.0\\ 
  
    \bottomrule
    \end{tabular}
    \caption{ The detailed Metaworld ML45 results of the IL-only, IL-only/single, SMART-only and SMART-only/single on each tasks.}
    \label{tab:task-detail-meta-2}
\end{table*}

\end{document}